\documentclass[acmlarge]{acmart}

\usepackage{graphicx}
\usepackage{subfigure}
\usepackage{multirow}
\usepackage{color}

\usepackage{mdframed}
\usepackage{enumitem}

\AtBeginDocument{%
  \providecommand\BibTeX{{%
    \normalfont B\kern-0.5em{\scshape i\kern-0.25em b}\kern-0.8em\TeX}}}

\author{Ziqi Gao}
\authornote{Both authors contributed equally to this research.}
\email{gzq22@mails.tsinghua.edu.cn}
\orcid{0009-0008-4553-7828}
\affiliation{%
  \institution{Key Laboratory of Pervasive Computing, Ministry of Education, Department of Computer Science and
Technology, Global Innovation Exchange (GIX) Institute, Tsinghua University}
  \city{Beijing}
  \country{China}
  \postcode{100084}
}
 
\author{Yuntao wang}
\orcid{0000-0002-4249-8893}
\email{yuntaowang@tsinghua.edu.cn}
\authornotemark[1]
\authornote{Corresponding author.}
\affiliation{%
  \institution{Key Laboratory of Pervasive Computing, Ministry of Education, Department of Computer Science and Technology, Tsinghua University}
  \city{Beijing}
  \country{China}
  \postcode{100084}
  }
\affiliation{%
  \institution{Department of Computer Technology and Application, Qinghai University}
  \city{Xining}
  \state{Qinghai}
  \country{China}
  \postcode{810016}
  }
 
\author{Jianguo Chen}
\orcid{0009-0003-4616-7500}
\email{jc2hk@virginia.edu}
\affiliation{
      \institution{University of Virginia}
      \city{Charlottesville}
      \state{VA}
      \country{USA}
}

\author{Junliang Xing}
\orcid{0000-0001-6801-0510}
\email{jlxing@tsinghua.edu.cn}
\affiliation{%
  \institution{Department of Computer Science and Technology, Tsinghua University}
  \city{Beijing}
  \country{China}
  }

\author{Shwetak Patel}
\orcid{0000-0002-6300-4389}
\email{shwetak@cs.washington.edu}
\affiliation{%
    \institution{Paul G. Allen School for Computer Science and Engineering, University of Washington}
    \city{Seattle}
    \state{WA}
    \country{USA}
    }

\author{Xin Liu}
\orcid{0000-0002-9279-5386}
\email{xliu0@cs.washington.edu}
\affiliation{%
    \institution{Paul G. Allen School for Computer Science and Engineering, University of Washington}
    \city{Seattle}
    \state{WA}
    \country{USA}
    }

\author{Yuanchun Shi}
\orcid{0000-0003-2273-6927}
\email{shiyc@tsinghua.edu.cn}
\affiliation{%
  \institution{Department of Computer Science and Technology, Tsinghua University}
  \city{Beijing}
  \country{China}
  \postcode{100084}}
\affiliation{%
  \institution{Qinghai University}
  \city{Xining}
  \state{Qinghai}
  \country{China}
  \postcode{810016}}

\renewcommand{\shortauthors}{Gao and Wang, et al.}
\setcopyright{rightsretained} 
\acmJournal{IMWUT}
\acmYear{2023} \acmVolume{7} \acmNumber{3} \acmArticle{96} \acmMonth{9} \acmPrice{}\acmDOI{10.1145/3610872}

\begin{document}



\title{MMTSA: Multi-Modal Temporal Segment Attention Network for Efficient Human Activity Recognition}





\begin{abstract}
  Multimodal sensors provide complementary information to develop accurate machine-learning methods for human activity recognition (HAR), but introduce significantly higher computational load, which reduces efficiency. 
  This paper proposes an efficient multimodal neural architecture for HAR using an RGB camera and inertial measurement units (IMUs) called Multimodal Temporal Segment Attention Network (MMTSA). 
  MMTSA first transforms IMU sensor data into a temporal and structure-preserving gray-scale image using the Gramian Angular Field (GAF), representing the inherent properties of human activities.
 MMTSA then applies a multimodal sparse sampling method to reduce data redundancy. 
Lastly, MMTSA adopts an inter-segment attention module for efficient multimodal fusion. 
Using three well-established public datasets, we evaluated MMTSA's effectiveness and efficiency in HAR.
Results show that our method achieves superior performance improvements ($11.13\%$ of cross-subject F1-score on the MMAct dataset) than the previous state-of-the-art (SOTA) methods. 
The ablation study and analysis suggest that MMTSA's effectiveness in fusing multimodal data for accurate HAR. 
The efficiency evaluation on an edge device showed that MMTSA achieved significantly better accuracy, lower computational load, and lower inference latency than SOTA methods. 
\end{abstract}





\maketitle

\section{Introduction}


The intersection of wearable sensors and deep learning has recently spurred interest in human activity recognition (HAR) within fields such as human-computer interaction (HCI), ubiquitous computing, and healthcare. The ability to recognize human activities provides computers with valuable insights into user behavior and intentions, leading to more natural human-computer interactions and context-aware capabilities. Accordingly, the demand for HAR systems that accurately recognize human activities while efficiently operating on edge devices (such as wearables and smartphones) has become increasingly prominent.  

Most existing HAR methods, which rely on a single modality such as RGB video, audio, acceleration, or infrared sequences, have been demonstrated to be able to recognize human activities~\cite{TSN,lin2019tsm,garcia2017similarity,slim2019survey,akula2018deep}. However, these unimodal methods are insufficient to achieve high accuracy in complicated scenarios with fine-grained human activities \cite{chen2021deep}. Inertial measurement units (IMUs) and RGB cameras are commonly used sensors due to their prevalence in daily life. However, video-based HAR methods are highly susceptible to illumination intensity, visual occlusion, and complex backgrounds, while IMU-based HAR methods are negatively affected by noisy or missing data and motion variance between users.
Leveraging both vision-based and IMU-based modalities is essential in cases where a single modality exhibits weaknesses, as it can improve the performance of human activity recognition (HAR) in multimodal ways. For example, distinguishing between eating and drinking activities based solely on IMU sensor data can be challenging due to the similarity of hand movement trajectories. However, the visual modality can be used to differentiate between the two activities based on the visual characteristics of the objects held by the hands. 


Recently, several multimodal deep-learning-based HAR methods have been proposed to enhance the recognition performance \cite{islam2022mumu,islam2020hamlet,islam2021multi}. 
However, most existing methods utilize dense sampling and heterogeneous sub-networks to extract unimodal features and fuse them at the end, with unsatisfying performance regarding accuracy, latency, and computational load. Specifically, existing HAR deep-learning approaches have the following drawbacks:

1) \textbf{Structure Divergence \& Loss of Temporal Correlation.} Owing to data heterogeneity, most existing methods feed unimodal data into separate sub-networks with different structures to extract features and fuse them at the end stages. This approach leads to a significant structure divergence between the IMU sensor and vision sensor data. Since IMU sensor data are one-dimensional time-series signals, most of the previous works utilized 1D-CNN, RNN or LSTM network to extract spatial and temporal features of raw IMU sensor data~\cite{steven2018feature,panwar2017cnn,wang2019human}. The vision sensor data of human activities, however, usually comprises images or videos with two or more dimensions, making it suitable for 2D-CNN or 3D-CNN to extract visual features \cite{simonyan2014two,karpathy2014large,sun2017lattice}.
The input form of existing multimodal learning models ignores the temporal synchronization correlation between multimodal data and loses valuable complementary information. 
Additionally, adding a new modality input to an existing model requires the design of a new sub-network specific to that modality, which can limit the model's generalization to new modalities. 

2) \textbf{Redundancy in Dense Sampling.} Dense temporal sampling, which involves sampling frames densely in a video clip or sampling the entire series of sensor data in a period, is widely used in previous work to capture long-range temporal information in long-lasting activities. However, those methods \cite{8964371,wei2019fusion} mainly rely on dense temporal sampling to improve the performance, which results in data redundancy and unnecessary computation since the adjacent frames in the video have negligible differences. Similarly, the IMU data of some activities (e.g., running, cycling) are periodic. Taking the whole IMU data of these activities as inputs reduces inference efficiency.

3) \textbf{Deployment Challenges due to Complexity.} Although some newly proposed attention-based multimodal learning methods have improved the performance of HAR tasks, their complicated architectures lead to high computational overhead and make them challenging to be deployed on mobile and wearable devices~\cite{islam2021multi,islam2022mumu}.

To address these challenges mentioned above, we propose \textbf{MMTSA}, a novel \textbf{M}ulti-\textbf{M}odal \textbf{T}emporal \textbf{S}egment \textbf{A}ttention neural architecture based on RGB camera and IMU sensor data for end-to-end human activity recognition application. We first utilize Gramian Angular Field (GAF) as a multimodal data isomorphism mechanism to represent the inherent properties of human activities in the IMU data. Then we apply a multimodal sparse sampling method to reduce data redundancy.  Lastly, we adopt an inter-segment attention module for efficient multimodal fusion at the end of MMTSA.
Using three well-established public datasets including MMAct~\cite{Kong_2019_ICCV}, DataEgo~\cite{possas2018egocentric} and Multimodal Egocentric Activity~\cite{song2014activity}, we evaluated MMTSA's effectiveness and efficiency in recognizing human activities.
The main contributions of this paper are summarized as follows:
\begin{itemize}

\item We propose a novel architecture called \textbf{MMTSA} for efficient human activity recognition. MMTSA adopts 2D-CNN as the backbone network, which utilizes multimodal data isomorphism mechanism based on Gramian Angular Field (GAF) IMU data imaging, segment-based multimodal sparse sampling, and inter-segment attention for efficient human activity inference.  
\item We evaluated and compared MMTSA's performance with SOTA methods on three public multimodal HAR datasets. Results show that MMTSA achieves an improvement of 11.13\% and 2.59\% (F1-score) on the MMAct dataset for the cross-subject and cross-session evaluations, respectively. The edge deployment evaluation shows that MMTSA achieves 16.2\% higher accuracy and reduces 94\% of FLOPs and 82.6\% of inference latency when compared with SOTA methods.
\item  
We thoroughly discuss key features of raw IMU data for HAR and analyze the characteristics of grayscale images derived from the GAF-based method, which elucidates the underlying physical properties of real-world IMU signals.
We demonstrate and analyze the reasons regarding the effectiveness of each  component within MMTSA through a series of ablation studies. 
\end{itemize}
\section{Related Work} 
\subsection{Unimodal Human Activity Recognition}
Unimodal HAR, which focuses on using a single modality (visual or wearable sensors) for activity recognition, has been extensively investigated in recent years. This section discusses recent research progress on unimodal HAR based on visual and IMU-sensor modalities.

\subsubsection{Video-based Human Activity Recognition}
Video-based HAR, in particular, has attracted significant attention due to the increasing availability of video data in various fields such as health care, sports analysis, and video surveillance. Deep learning methods for video-based HAR have gained widespread popularity in recent years due to their superior performance and ability to learn complex representations from raw data automatically. One major trend is the use of deep learning methods such as convolutional neural networks (CNNs), recurrent neural networks (RNNs), and long short-term memory (LSTM) networks. \citet{simonyan2014two,feichtenhofer2016convolutional,chi2019two} used two-stream CNN networks to incorporate spatial and temporal information from RGB frames and optical flow. Unlike traditional 2D CNNs or two-stream CNNs, \citet{chenarlogh2019multi} proposed a multi-stream 3D-CNN network to capture better the dynamic spatiotemporal information of human activities in videos. C3D \cite{DuTrancnnvideo2015}, I3D \cite{carreira2017quo} used 3D convolutional kernels to better process both spatial and temporal information from video data, outperformed traditional 2D CNNs or two-stream models. \citet{TCN, TSN, lin2019tsm, feichtenhofer2019slowfast} focused on temporal modeling, which allowed for modeling both short-term and long-term temporal dependencies in videos to improve the recognition performance. Vision transformers (ViT) \cite{vit} has been widely applied in video-based HAR tasks due to their improved representation learning. RViT \cite{yang2022recurring} proposed a novel Recurrent Vision Transformer that can capture spatiotemporal features via the attention gate and recurrent execution and can support variant-length videos as inputs. TimeSformer \cite{bertasius2021space} proposed a divided space-time attention mechanism for vision transformer, reducing the amount of calculation and improving the recognition precision.


Although deep learning methods for video-based HAR have good performance in many scenarios, they face several challenges, including high computational complexity, limited ability to capture the temporal dynamics of human activities, and high sensitivity to visual context (light intensity, occlusion, viewing angle).

\subsubsection{IMU-based Human Activity Recognition}
IMUs provide continuous, non-intrusive, and cost-effective monitoring of human activities or movements. Several deep learning or rule-based methods have been proposed for IMU-based HAR or body movement~\cite{14convolutional,wristcnn,ignatov2018real, chi17-float, imwut-dualring, chi22-faceori}. \citet{steven2018feature, DRNN, wang2019human} proposed several LSTM or RNN-based HAR models to extract spatial and temporal features from raw IMU signals. \citet{lu2019robust,wang2015ensorencoding} utilized math tools to transform the IMU time series into color images so that 2D-CNNs can applied. 
\citet{tong2021zero} proposed a zero-shot learning method for IMU-based HAR. In this approach, the pre-trained video embeddings are used to augment the IMU data and provide auxiliary information, which helps the IMU-based HAR model to recognize unseen activities. However, the accuracy of IMU sensor-based methods is sensitive to the placement on the human body \cite{mukhopadhyay2014wearable}. Furthermore, most of the current IMU sensor-based methods perform poorly in complex HAR scenarios.      

Although these single-modality methods have shown promising performances in many cases, these approaches have a significant weakness: they rely on high-quality sensor data. If the single-modality data is noisy and missing, the unimodal learning methods cannot extract robust features and perform poorly in human activity recognition. 

\subsection{Multimodal Human Activity Recognition}
To overcome the shortcoming of single modality missing and occlusion, multimodal learning methods have been used in HAR. By aggregating the advantages and capabilities of various data modalities, multimodal learning can provide more robust and accurate HAR. Thus, learning outstanding multimodal features is a critical challenge in designing a powerful multimodal feature learning approach. Several approaches \cite{joze2020mmtm, liu2021semantics, tong2021zero, chi22-faceori, chi23-earcough} have been proposed to fuse these sensor data from different modalities. For each modality, a domain-specific feature encoder sub-network is used to extract feature representations, and then all modalities' feature representations are concatenated at the end of the framework for classification. Therefore, the final performance is highly related to salient feature representations of a single modality. However, these architectures neglect the intrinsic synchronous property among all modalities and assume all modalities contribute to final performance equally.

To address these challenges, several works introduce new multimodal HAR algorithms. Firstly, multi-task and multi-stage deep learning methods have been used to design a new framework that learns to combine features from different sensor modalities effectively. MuMu~\cite{islam2022mumu} proposed a cooperative multitask learning scheme by creating an auxiliary task and a target task, where the auxiliary task guided the target task to extract complementary multimodal representations appropriately. Additionally, \citet{choi2022multi} proposed a two-stage feature fusion method, wherein the first stage, each input encoder learned to extract features effectively and in the second stage, learned to combine these individual features. Secondly, instead of aggregating different modalities late, TBN~\cite{kazakos2019epic} combined three modalities (RGB, flow, and audio) with mid-level fusion at each time step. It showed that visual-audio modality fusion in egocentric action recognition tasks improved the performance of the action and accompanying object. However, the mid-level fusion method is only explored in video and audio modalities and has not been extended to other sensor data (e.g., IMU sensors). Furthermore, attention-based approaches have recently been applied in feature learning for HAR. The attention mechanism allows the feature encoder to focus on specific parts of the representation while extracting the salient features of different modalities. For example, \citet{long2018multimodal} proposed a new kind of attention method called keyless to extract salient unimodal features combined to produce multimodal features for video recognition. HAMLET \cite{islam2020hamlet} proposed a hierarchical multimodal attention method for extracting salient unimodal features and fusing those to generate multimodal features for HAR. Moreover, Multi-GAT \cite{islam2021multi} explored the possibilities of using graphical attention methods for multimodal representation learning in HAR. 

Although these multimodal HAR methods have achieved good performance in various scenarios, several challenges still remain in multimodal HAR. For example, many of these methods encode the whole sensor data, which is redundant and highly computational. Therefore, a sparse and efficient sampling strategy would be more favorable and need to be designed. Furthermore, many existing frameworks do not allow inter-modality interaction, which can cause potential loss of inter-modality correlation and may not learn complementary multimodal features. Thus, we explore the inter-segment modality attention mechanism and demonstrate that it improves the final result. Finally, we propose a novel multimodal temporal segment attention network MMTSA, as described in detail in the next section.

\section{Motivation} 
In this section, we observe and explore the characteristics of multimodal raw data and the challenges of data processing in human daily activities. We analyze the shortcomings of existing research methods and illustrate the strengths of MMTSA in addressing existing challenges.

\subsection{Observations on the Properties of IMU Data in Human Activities} \label{IMU-moti}

\begin{figure}
	\centering
	\subfigure[Running]{
		\begin{minipage}[b]{0.43\linewidth}
            \centering
			\includegraphics[width=1\linewidth]{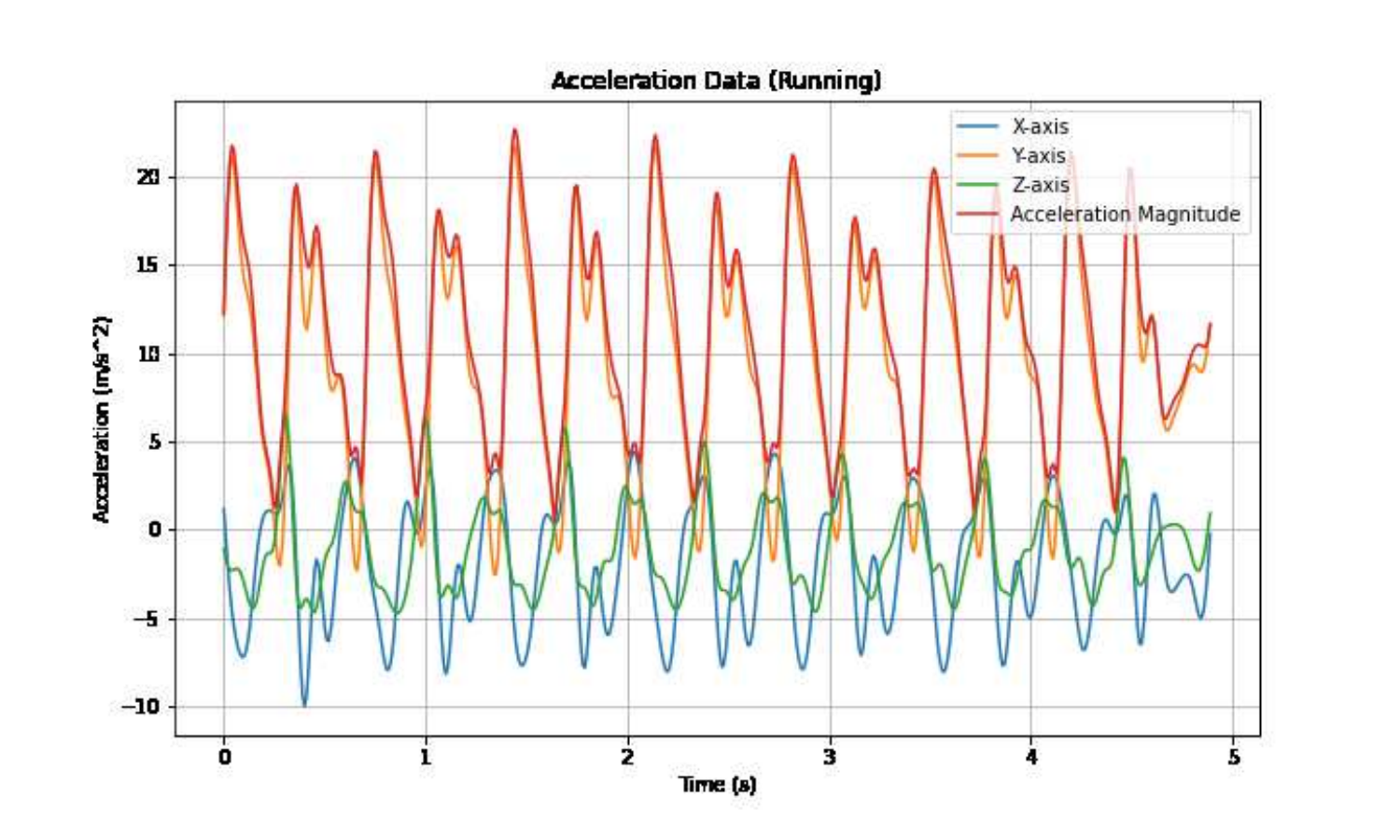} 
		\end{minipage}
		\label{run}
  }
    \subfigure[Walking]{
        \begin{minipage}[b]{0.43\textwidth}
            \centering
            \includegraphics[width=1\textwidth]{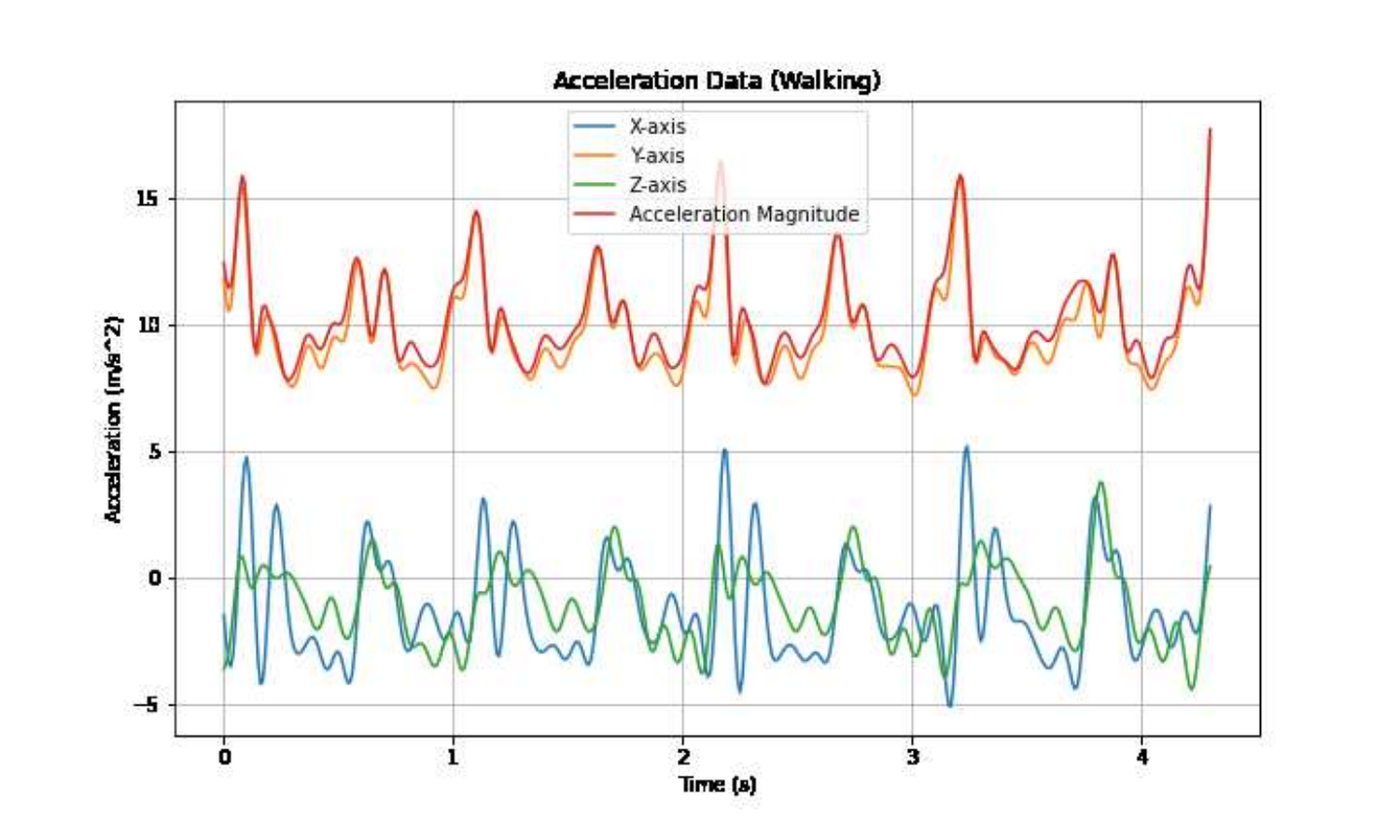}
        \end{minipage}
        \label{walk}
        }
        \\
    \subfigure[Opening]{
        \begin{minipage}[b]{0.43\textwidth}
            \centering
            \includegraphics[width=1\textwidth]{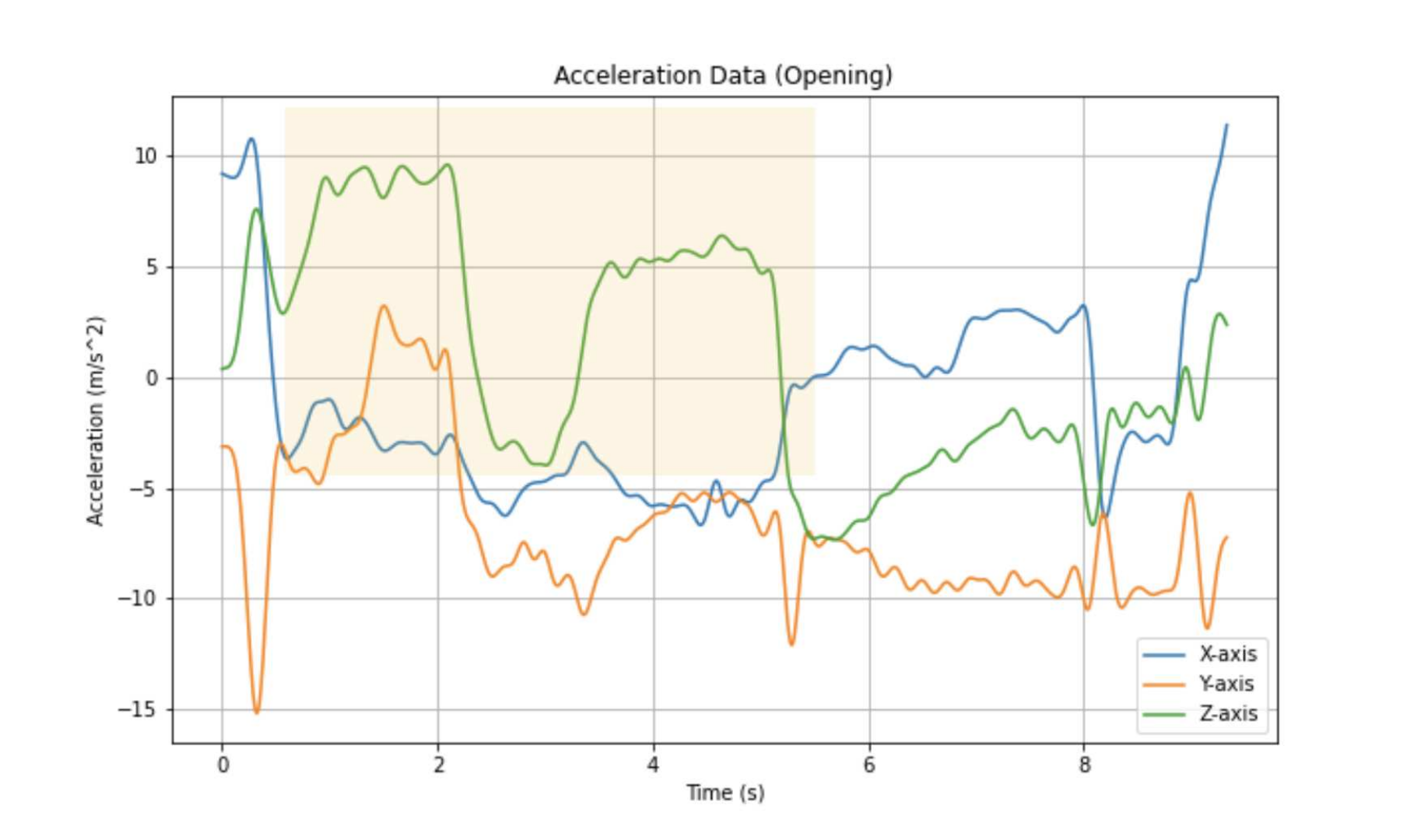} 
        \end{minipage}
        \label{open}
        }
    \subfigure[Closing]{
        \begin{minipage}[b]{0.43\textwidth}
            \centering
            \includegraphics[width=1\textwidth]{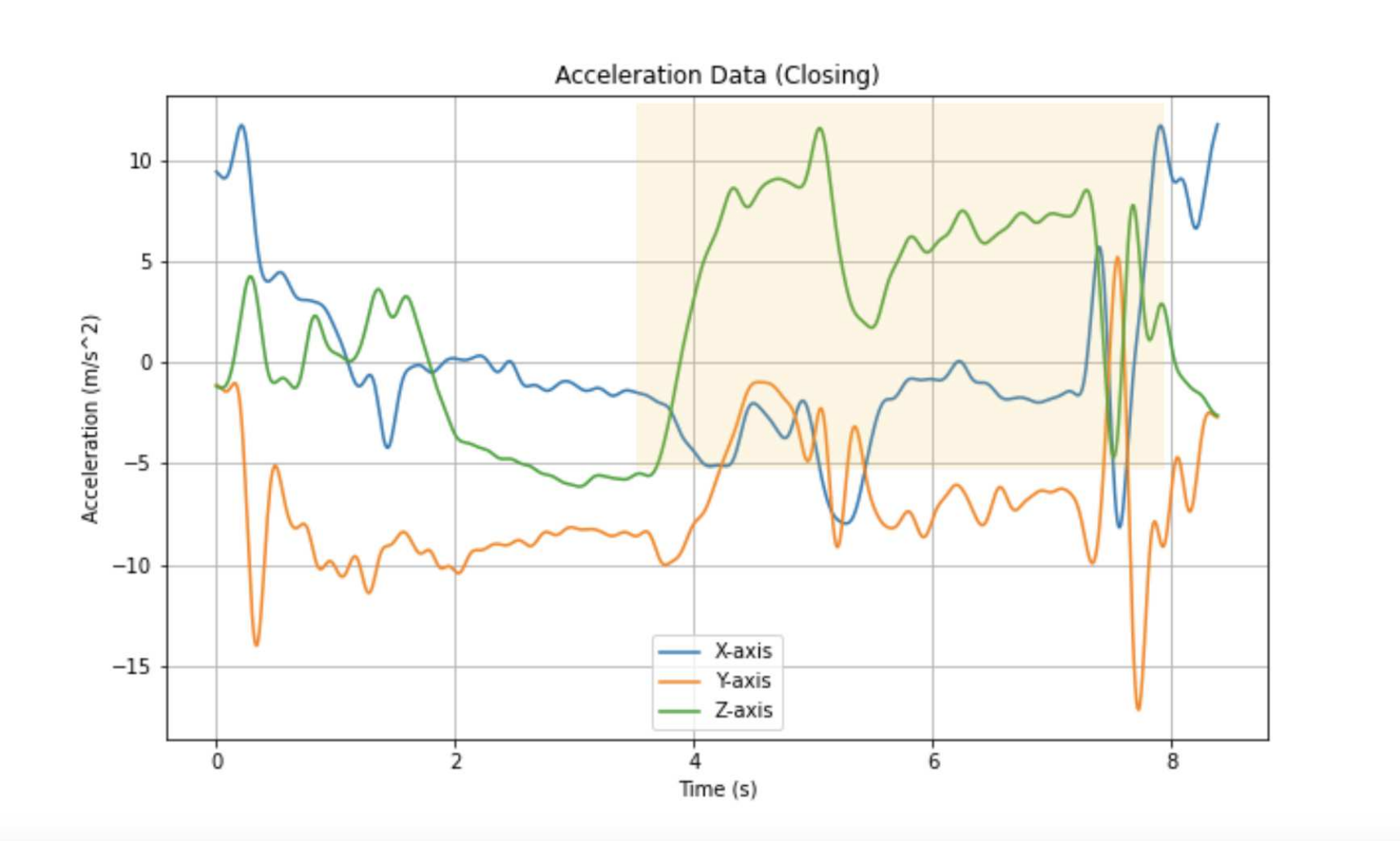} 
        \end{minipage}
        \label{close}
        }
	\caption{Signals of the accelerometer in the smartphone when Subject 12 from the MMAct dataset performs four activities. Compared to Opening and Closing, significant periodicity in the IMU signal can be observed during walking and running. The peak amplitude and cycle length of the IMU signals can be used to distinguish between running and walking, which represent the subject's instantaneous acceleration and cadence, respectively. Similar local waveforms appear in the Z-axis components of the IMU signals for opening and closing (yellow areas), but their sampling timestamps are different, which indicates that the sub-action sequences of the two activities are different.}
	\label{fig-acclero}
\end{figure}
In this section, we delve into the properties of IMU (Inertial Measurement Unit) sensor data collected from human activity datasets in real-world environments. We analyze two representative datasets: MMAct \cite{Kong_2019_ICCV} and DataEgo \cite{possas2018egocentric}, which provide insights into human activities using various wearable devices, such as smartwatches, smartphones, and AR glasses.

Intuitively, a person's daily activities are often comprised of fine-grained actions over a period of time. To distinguish different activities, we need to consider physical characteristics such as the order, periodicity, amplitude, and limb movement patterns of fine-grained actions. For instance, the arm swing period during running is shorter and has a larger amplitude compared to that of walking, and the trajectory of arm movement also differs between the two activities. 


We observe and analyze how these physical features that distinguish daily activities are implied in IMU data. Figure \ref{fig-acclero} illustrates the accelerometer data recorded from a smartphone worn by Subject 12 in the MMAct dataset. The data captures four activities: running, walking, opening, and closing. 

Representing the periodic waveform of IMU data helps to distinguish prolonged periodic activities (eg, running, walking) from other non-periodic activities. By examining the waveform diagram of the accelerometer's y-axis component, we observe significant periodicity in the IMU signal during walking or running and the peaks of the signal align with the forward steps of the subject. This periodicity is absent from the waveforms corresponding to the other two activities.

Preserving the original data sampling and temporal information while modeling IMU data is essential for accurate activity recognition. 
Although accelerometer data for running and walking (Figures \ref{run} and \ref{walk}) share similar waveforms and periodicity in the y-axis component, they can be distinguished based on the peak amplitude and cycle length. The peak amplitude indicates the subject's stride frequency and instantaneous acceleration, while the cycle length represents the speed of the stride frequency. 
Preserving the real sampling timestamp is useful to indicate the order of fine-grained actions, which is beneficial to model IMU time series and distinguish coarse-grained activities. For example, opening and closing a locker (Figures \ref{open} and \ref{close}) include the two sub-actions of unlocking and opening/closing the cabinet door. Examining the accelerometer signal recorded from a smartwatch worn by Subject 12, we observe a similarity in the z-axis component of the signal during both activities. This similarity arises from the similar vertical arm movement when picking up the key to open or close the lock. Ignoring the timestamp information and solely modeling the sampling data would make it challenging to differentiate between these activities. However, the order of sub-actions, such as unlocking the lock before opening the door, is reflected in the timestamp of signal sampling at different stages. 

\begin{figure}[htp]
    \centering
    \includegraphics[width=13cm]{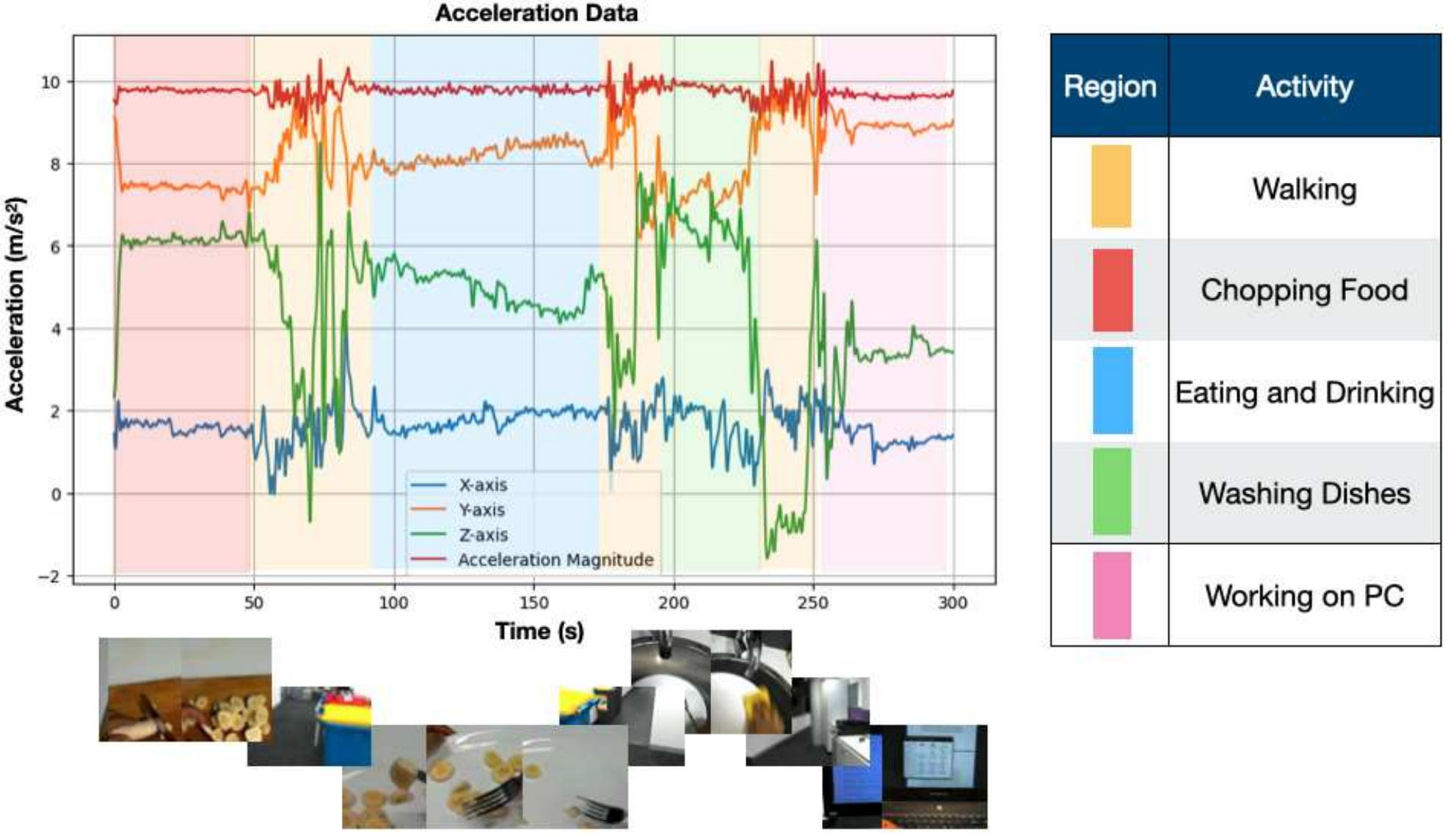}
    \caption{Accelerometer data and its synchronized RGB video frames collected by subjects in the Dataego dataset during daily continuous activities. Fluctuations in the accelerometer signal are notably observed during activity transitions and show periodicity when the subject's position is relatively static, such as in dishwashing.}
    \label{fig:ego-acc}
\end{figure}

The DataEgo dataset provides additional insights into continuous daily activities using smart AR glasses. Analyzing the first-person video and accelerometer data captured during these activities (Figure \ref{fig:ego-acc}), we face challenges such as relatively static head postures during some activities (e.g., reading or working on a PC). This leads to limited information contained in the data. To tackle these challenges, we can extract features from critical moments as well as individual cycle waveforms of periodic IMU data series, enhancing human activity modeling in scenarios with less informative or noisier sensor data. We observe that significant fluctuations occur in the accelerometer signal when the subject initiates or concludes an activity, particularly during transitions between dynamic and static activities. Additionally, the waveform of the accelerometer exhibits periodicity during activities where the subject's position remains relatively fixed, such as washing dishes. 

In summary, an ideal modeling approach for IMU sensor signals in human activity recognition should incorporate the following elements: (1) retaining real sampling information such as intensity and amplitude, (2) maintaining timestamp information, (3) effectively extracting temporal and spatial features to recognize waveform and periodicity while reducing noise and irrelevant information, and (4) accurately identifying signal changes indicating the start or end of activities for long-term daily activity recognition.

However, existing methods that utilize IMU signals in human activity recognition have limitations to incorporate the above elements. Statistical-based methods, which typically calculate statistical measures, such as mean, variance, or correlation coefficients, often overlook temporal dynamics and fail to capture the waveform features and periodicity. Spectrogram-based methods suffer from information loss due to the transformation process, they often focus solely on frequency information and overlook the temporal and spatial characteristics of the signals. Traditional machine learning methods struggle to handle complex temporal dependencies and may not fully leverage the temporal and spatial information present in the IMU data. Additionally, their performance is sensitive to manual feature engineering. 1D-LSTM and CNN can capture temporal dependencies and learn hierarchical representations from the IMU signals. However, they still face challenges in effectively utilizing the intensity and amplitude information, preserving the timestamp information, and extracting long-term temporal correlations without distortion. 

Unlike any of the above methods, MMTSA innovatively uses the imaging mechanism to map the IMU signal to a high-dimensional space, which not only reduces the structural differences of different modal data, but also satisfies the elements of the ideal HAR modeling approach mentioned above. See details in Section \ref{gaf_image}.
 
\subsection{Redundancy of Data from Different Modalities in Human Activities} \label{redu}
The redundancy present in different modalities of data used for human activity recognition is observed and analyzed in this section. Visual and IMU sensors continuously capture data while individuals engage in their daily activities. Figure \ref{fig:ego-acc} illustrates the video data and three-axis acceleration data recorded by the sensor on a subject's AR glasses in the DataEgo dataset during various daily activities over a 5-minute period.

Regarding the visual modality, the high sampling rate of the camera results in minimal pixel differences between frames of a video. Consequently, the amount of new information provided in each subsequent frame is relatively low compared to the previous frame. This leads to redundancy and duplication of information captured by the vision-based sensor. Furthermore, our analysis reveals that when subjects perform activities with a relatively fixed position, the visual sensor records even more redundant information. For instance, Figure \ref{fig:ego-acc} displays multiple frames from the video where the subject is eating. Although fine-grained information, such as the number of banana slices on a plate, varies between frames, the key objects (forks, food, and plates) appearing in the frames remain the same. This coarse-grained information, which is crucial for identifying the subject's current activities, is already captured in some key frames of the video data. Therefore, these key frames contain most of the necessary information for identifying the entire activity.

Similarly, redundancy also exists in the IMU-based modality information. In Figure \ref{fig:ego-acc}, the y-axis and z-axis components of the watch's accelerometer exhibit periodic patterns when the subject is washing dishes. This phenomenon is more pronounced in activities involving regular limb movements, such as running or walking, as shown in Figure \ref{run} and \ref{walk}. As discussed in Section \ref{IMU-moti}, the periodicity of the IMU sensor signal waveform reveals specific movement patterns of body parts during these activities. Therefore, by effectively extracting finer-grained waveform features from a segment of IMU signals, we can model human behavior during these activities.

Based on the aforementioned analysis, it is evident that when modeling multimodal sensor data for human activity recognition, a sparse sampling strategy should be employed to avoid using all the original data as input. This approach, which is used in MMTSA (Section \ref{sbmss}), serves to reduce the inclusion of redundant information, accelerate the inference speed of the model, and enhance its real-time and efficient recognition capabilities.

\section{Method}

\begin{figure}[htp]
    \centering
    \includegraphics[width=13cm]{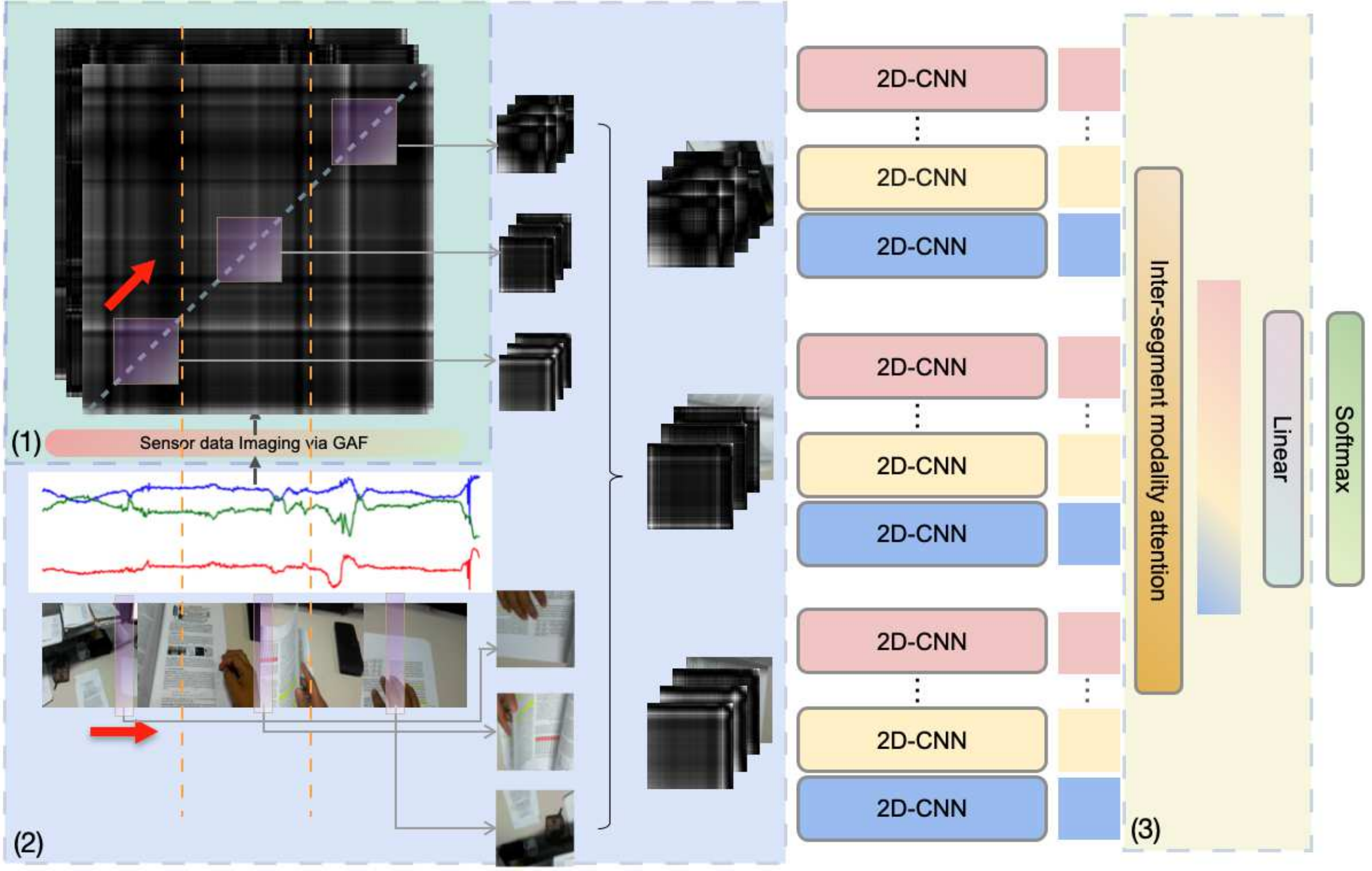}
    \caption{The architecture of MMTSA: (1) Multimodal data isomorphism mechanism based on GAF. (2) Segment-based multimodal sparse sampling. (3)Inter-segment attention modality fusing.}
    \label{fig:Architecture}
\end{figure}

In this section, we present our proposed method: MMTSA, a multimodal temporal segment attention network as shown in Fig. \ref{fig:Architecture}. MMTSA consists of three sequential learning modules: 
\begin{itemize}
    \item \textbf{Multimodal data isomorphism mechanism based on IMU data imaging}: The module is responsible for transforming IMU sensor data into multi-channel grayscale images via Gramian Angular Field (GAF), making visual-sensor data and IMU sensor data representations to be isomorphic.
    \item \textbf{Segment-based multimodal sparse sampling}: We propose a novel multimodal sparse sampling strategy in this module. It performs segmentation and random sampling on RGB frame sequences and GAF images of IMU data, preserving the modal timing correlation while effectively reducing data redundancy.
    \item \textbf{Inter-segment attention for multimodal fusing}: To better mine the spatiotemporal correlation and complementary information between modalities, we propose an efficient inter-segment attention method to fuse multimodal features, which improves HAR performance. 
\end{itemize}
We discuss how MMTSA works in greater detail.

\subsection{Multimodal Data Isomorphism Mechanism} \label{gaf_image}
Although deep learning has achieved great success in CV and NLP, its techniques fail to have many comparable developments for time series. Most traditional deep learning methods for time series build models based on RNN, LSTM, or 1D-CNN. However, these approaches have been proven to have limitations \cite{pascanu2013difficulty,wu2018comparison}. 

To leverage the advanced 2D-CNNs and related techniques in computer vision, \citet{wang2015ensorencoding} first proposed a novel representation for encoding time series data as images via the Gramian Angular Fields (GAF) and hence used 2D-CNNs to improve classification and imputation performance. Since then, time series imaging methods have caught much attention. Inspired by the method proposed in \cite{wang2015ensorencoding}, we note that GAF-based methods have great potential to reduce structural differences in data from multiple modalities (e.g., RGB video and accelerometers). Therefore, we propose a multimodal data isomorphism mechanism based on GAF, which can enhance the representation ability of the temporal correlation and inherent properties of IMU sensor data and improve the reusability of different modal feature extraction networks, see Figure \ref{fig:GAF}. We will now briefly describe how our multimodal data isomorphism mechanism works. 

\subsubsection{IMU Sensor Series Rescaling}
Let $S=\left\{s_{t_1}, s_{t_2}, \ldots, s_{t_n}\right\}$ be a time series collected by an IMU sensor, where $s_{t_i} \in \mathbb{R}$ represents the sampled value at time $t_i$. $T = t_n - t_1$ represents the sampling time duration of this time series. We rescale $S$ onto $[-1,1]$ by:

\begin{equation}
\begin{aligned}
& \tilde{s_{t_i}} &=\frac{\left(s_{t_i}-\max (S)+\left(s_{t_i}-\min (S)\right)\right.}{\max (S)-\min (S)}.
\label{eqn:scale}
\end{aligned}
\end{equation}
The max-min normalization step makes all values of  $S$ fall in the definition domain of the \textit{arccos} function, which satisfies the conditions for the coordinate system transformation.

\subsubsection{Polar Coordinate System Transformation}
In this step, we transform the normalized Cartesian IMU sensor series into a polar coordinate system. For the time series $S$, the timestamp and the value of each sampled data point need to be considered during the coordinate transformation. Then we use an inverse cosine function to encode each data point $\tilde{s_{t_i}}$ into polar coordinate by:

\begin{equation}
\left\{\begin{array}{c}
\phi_{t_i}=\arccos \left(\tilde{s_{t_i}}\right),-1 \leq \tilde{s_{t_i}} \leq 1, \tilde{s_{t_i}} \in \tilde{S} \\
r_{t_i}=\frac{t_i}{T}, t_i \in \mathbb{T}
\end{array}\right.,
\label{eqn:polar}
\end{equation}
where $\phi_{t_i}$ and $r_{t_i}$ represent the angle and the radius of $\tilde{s_{t_i}}$ in the polar coordinate, respectively. The encoding in equation\ref{eqn:polar} has the following advantages. First, it is a composition of bijective functions as $\cos \left(\phi\right)$ is a monotonic function when $\phi \in [0, \pi]$, which allows this transformation to preserve the integrity of the original data. Second, it preserves absolute temporal relations, as the area of $\phi_{t_i}$ and $\phi_{t_j}$ in polar coordinates is dependent on not only the time interval of $t_i$ and $t_j$, but also the absolute value of them \cite{wang2015ensorencoding}. The coordinate transformation above maps the 1D time series into a 2D space, which is imperative for later calculating the Gramian Angular Field.
\begin{figure}[htp]
    \centering
    \includegraphics[width=0.8\linewidth]{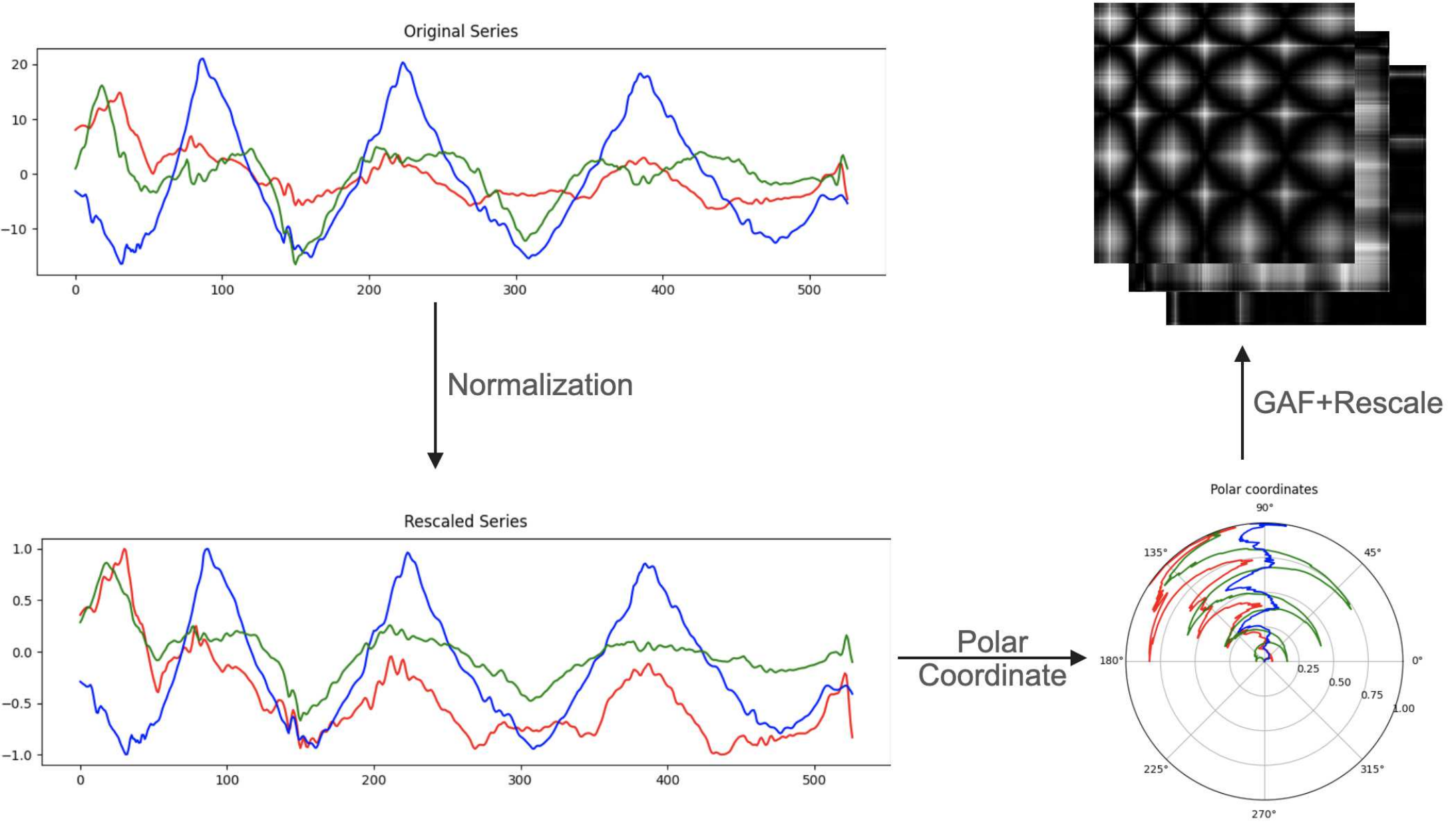}
    \caption{IMU sensor data imaging via Gramian Angular Field in MMTSA. The IMU data in the figure is sampled from the three-axis accelerometer of the subject in the MMAct dataset while shaking hands.The IMU time series is first rescaled onto $[-1,1]$, and then transformed from Cartesian to polar coordinates. The IMU time series in a polar coordinate is represented by Gramian Angular Field and mapped to grayscale images.}
    \label{fig:GAF}
\end{figure}

\subsubsection{Image Generation via Gramian Angular Field (GAF)} \label{sec:img-gaf}
The Gram Matrix G is the matrix of all inner products of $X=\left\{x_1, x_2, \ldots, x_n\right\}$, where $x_i$ is a vector. A dot product in a Gram Matrix can be seen calculating the similarity between two vectors. However, in the polar coordinate mentioned above, the norm of each vector causes a bias in the calculation of the inner product. Therefore, we exploit the angular perspective by considering the trigonometric sum between each point in the polar coordinate to identify the temporal correlation within different time intervals, which is an inner product-like operation solely depending on the angle. The GAF is defined as follows: 
\begin{equation}
\begin{aligned}
&G(\tilde{S})=\left[\begin{array}{ccc}
\cos \left(\phi_{t_1}+\phi_{t_1}\right) & \cdots & \cos \left(\phi_{t_1}+\phi_{t_n}\right) \\
\cos \left(\phi_{t_2}+\phi_{t_1}\right) & \cdots & \cos \left(\phi_{t_2}+\phi_{t_n}\right) \\
\vdots & \ddots & \vdots \\
\cos \left(\phi_{t_n}+\phi_{t_1}\right) & \cdots & \cos \left(\phi_{t_n}+\phi_{t_n}\right)
\end{array}\right],\\
\end{aligned}
\label{eqn:gaf}
\end{equation}
where
\begin{equation}
\begin{aligned}
\cos \left(\phi_{t_i}+\phi_{t_j}\right) =& \cos (\arccos (\tilde{s_{t_i}})+\arccos (\tilde{s_{t_j}})) \\
=& \tilde{s_{t_i}} \cdot \tilde{s_{t_j}} - \sqrt{1-\tilde{s_{t_i}}^2} \cdot \sqrt{1-\tilde{s_{t_j}}^2} , 1 \leq i,j \leq n.\\
\label{eqn:func}
\end{aligned}
\end{equation}
It should be emphasized that the GAF mentioned in this paper actually refers to the Gramian Angular Summation Field (GASF) in the cosine form. Another form of GAF is the Gramian Angular Difference Field (GADF) which uses the sine function to represent the difference between relative phases. Compared to GADF, GASF utilizes the cosine function to represent the sum of relative phases, more effectively capturing the periodicity and temporal correlation of time series. 
When there is a greater correlation between two sampled data points, the value computed by the GASF operator is also higher. In contrast, the GADF operator cannot represent this correlation as effectively. \citet{wang2015ensorencoding} also suggested using Markov Transition Fields (MTF) to visualize IMU data as images. However, compared to the GAF-based approach, the MTF method often involves estimating state transition probabilities, which are less intuitive and harder to interpret. Consequently, we choose the Gramian Angular Summation Field as the imaging method for the IMU time series in MMTSA.

The GAF representation in a polar coordinate maintains the relationship with the original time-series data via exact inverse operations. Moreover, the time dimension is encoded into GAF since time increases as the position moves from top-left to bottom-right, preserving temporal dependencies.

Then we map each element in the GAF representation to a pixel of a grayscale image by:
\begin{equation}
\tilde G_{i,j} = \frac {\cos(\phi_{t_i}+\phi_{t_j})-\min (\tilde G)} {\max (\tilde G) - \min (\tilde G)} \times 256,   1 \leq i,j \leq n,
\label{eqn:grey}
\end{equation}
where $\tilde G$ is a grayscale image of size $n \times n$.

Most wearable sensors (accelerometers, gyroscopes, magnetometers, etc.) are triaxial. Suppose given a time series data $\{S^{(x)}, S^{(y)}, S^{(z)}\}$ sampled by a 3-axis accelerometer, we can generate three grayscale images $\{\tilde G^{(x)}, \tilde G^{(y)}, \tilde G^{(z)}\}$ for the $x$, $y$, and $z$ axes according to the above steps. After concatenating these three images, the time series sampled by each three-axis IMU sensor can be uniquely converted to a multi-channel grayscale image of size $3 \times n \times n$.

\subsubsection{The Effectiveness Analysis of Imaging IMU Sensor Data based on GAF}

In this section, we analyze the imaging method of IMU sensor data based on GAF and explore the role and effectiveness of transforming IMU time series into a two-dimensional space for spatiotemporal feature extraction. As discussed in Section \ref{IMU-moti}, we have observed and analyzed IMU data recorded during actual human activities and presented several rules for effectively modeling IMU sensor signals. The GAF-based IMU data imaging method meets the requirements of these rules.

\begin{figure}[htp]
    \centering
    \includegraphics[width=13cm]{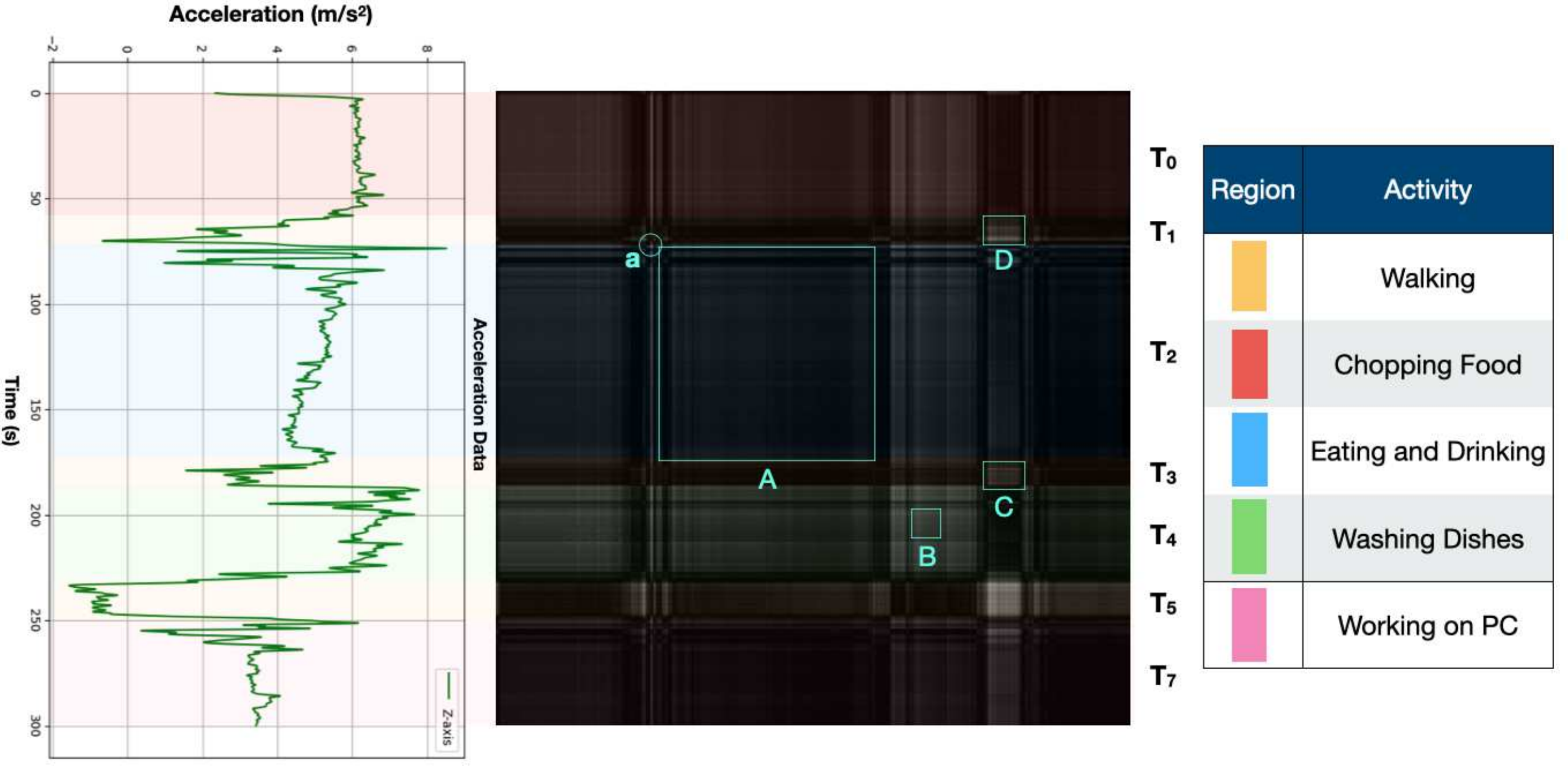}
    \caption{The z-axis component of the accelerometer and its corresponding GAF grayscale image of the 5-minute continuous activity in the DataEgo dataset. \textbf{Point a}, whose pixel value is much larger than other points around, is the beginning of the subject’s eating activity. \textbf{Area A}, which has an obvious boundary and texture, corresponds to the activity of eating food. \textbf{Area B} corresponds to the local waveform of the IMU signal while washing the dishes. This waveform pattern repeats during washing the dishes, which shows the obvious periodicity of this activity. \textbf{Area C} and \textbf{D} have relatively larger pixel values, which means the activities information between ${T_1}$ and $T_5$, $T_3$ and $T_5$ time periods are quite similar. }
    \label{fig:gaf-ana}
\end{figure}

Using accelerometer data from daily activities in Figure \ref{fig:ego-acc} as an example, we generate a GAF-based grayscale image of the z-axis component, as shown in Figure \ref{fig:gaf-ana}. The generated image is aligned with the accelerometer time series according to their timestamps. We mark some key points and areas in the image to analyze the GAF-based imaging method.

First, raw accelerometer data is rescaled onto the $[-1,1]$ range in Equation \ref{eqn:scale}. This normalization operation filters the data's overall bias while preserving the raw IMU signal's relative intensity magnitude and direction. This preserved information is essential for distinguishing signals with similar waveforms and identifying activity changes.

Equation \ref{eqn:func} denotes a correlation function used to compute the similarity between two samples of IMU data taken at different time points. The range of similarity values is $[-1, 1]$. The similarity is influenced by the direction and magnitude of the samples. When both samples exhibit the same direction (i.e., both positive or negative), a higher intensity in both signals corresponds to a greater similarity. Conversely, if the samples have opposite directions, a higher intensity leads to a lower similarity. By capturing the physical characteristics of the signal, this similarity function effectively captures the temporal correlation of IMU sensor data along the time dimension. The temporal correlation features imply the local waveforms and periodicity inherent in the raw IMU sensor data. Furthermore, when the two inputs represent signals sampled at the same time, the similarity function will degenerate into a quadratic function of a single variable,
\begin{equation}
G_{i,i} = 2 \tilde{s_{t_i}}^2-1, 1 \leq i \leq n,
\end{equation}
where $G_{i,j}$ will be proportional to the square of the sampled signal intensity.

Equations \ref{eqn:gaf} and \ref{eqn:grey} represent the value of each pixel in the grayscale image generated based on the similarity function. The diagonal of this pixel matrix preserves information about the original IMU signal's relative intensities and sampling timestamps. The higher the intensity of the sampled signal, the larger the value of the corresponding pixel on the diagonal line. The diagonal of the matrix goes from top left to bottom right with increasing timestamps. As we analyzed in Section \ref{IMU-moti}, the sharp increase or decrease in the signal intensity of the IMU sensor often corresponds to the moment when an activity starts or ends. Figure \ref{fig:gaf-ana} shows this evidence in the grayscale image. \textbf{Point a} has a much larger pixel value than other points around, which means the IMU signal's intensity at that moment increased sharply. In the synchronized video, the subject sat down at the moment corresponding to \textbf{Point a}. Before this, the subject was walking, after which they sat down and began to eat food. Therefore, the moment corresponding to \textbf{Point a} begins the subject's eating activity.

The square area along the diagonal of the grayscale image contains the time correlation information of the sampled signal within a period. For example, \textbf{Area A} is a pixel matrix of sampled data during the $T_2$ time period, corresponding to the subject's eating food activity. \textbf{Area A} and other areas of the grayscale image have apparent boundaries, and the texture in this area is also significantly different from other areas. This indicates that the GAF-based IMU signal imaging process extracts specific features of different activities, essential for recognizing motion patterns during various activities.

Observing the grayscale area corresponding to the subject washing dishes, we find that \textbf{Area B} corresponds to a local pattern of the IMU signal in this activity. This pattern is evenly distributed in the grayscale area of the dishwashing time range, which shows that the IMU sensor data of dishwashing behavior has evident periodicity. Thus, the IMU signal imaging method based on GAF can effectively extract this periodicity and local motion patterns.

The asymmetrical area outside the diagonal line in the grayscale image indicates the correlation between the IMU data collected in different periods. Suppose the pixels' value in this type of area is large. In that case, it means that during the two periods corresponding to the area, the similarity of IMU signals is high, and the activities of people during these two periods are more similar. \textbf{Area C} and \textbf{D} contain IMU signal's similarity information between ${T_1}$ and $T_5$, $T_3$ and $T_5$ time periods, respectively. We find that the pixel values in these two areas are relatively large, which means the subject's behavior is similar in these periods. This is consistent with the real situation, as the subjects walked during all three periods. This further illustrates the effectiveness of the GAF-based imaging method in extracting temporal correlations of IMU sensor data at different scales.

In conclusion, GAF-based imaging operation visually represents the IMU data's intensity and texture information while preserving its timestamp information and multi-scale temporal correlations. This can be useful for analyzing the waveform in the IMU data and identifying the underlying physical processes being measured. The generated grayscale images can reveal the start or end point of activity, waveform features, periodicity, and temporal correlation of the samples. These specific pieces of information are suitable as input for 2D convolutional neural networks for feature extraction and signal modeling.

\subsection{Segment-based Multimodal Sparse Sampling} \label{sbmss}

\begin{figure}[h]
    \centering
    \includegraphics[width=0.9\linewidth]{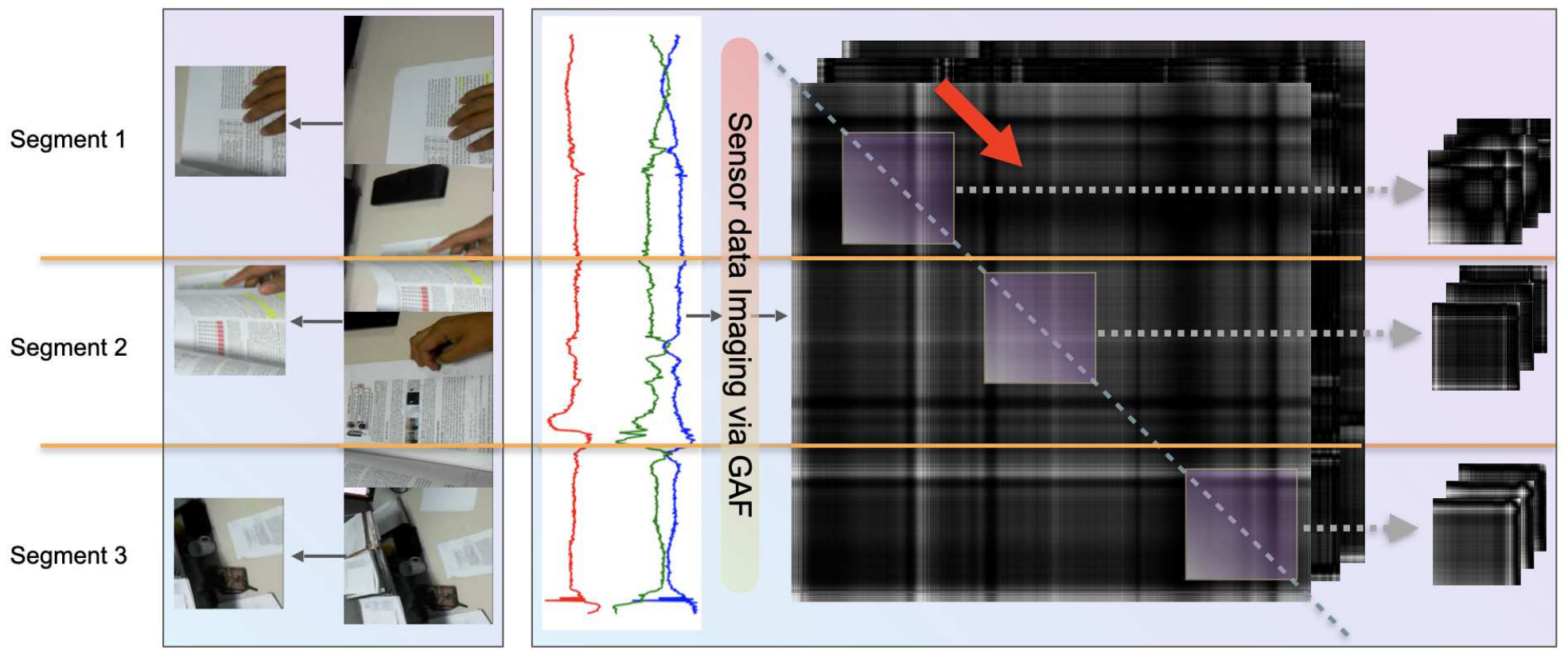}
    \caption{Multimodal sparse sampling strategy in MMTSA. A multimodal data clip is divided into N segments with the same duration, and a frame of the video is randomly selected in each segment as a snapshot. According to the timestamp of the snapshot, select a fixed-size square area on the diagonal of the grayscale image as a slice of the IMU data.}
    \label{fig:sparse_sample}
\end{figure}
The two mainstream methods of traditional human activity recognition based on video understanding are 3D-CNN and two-stream CNN networks. Still, the limitations of these two schemes are that they only can capture short-range temporal dependencies in the video. These methods usually require densely sampled video clips to capture long-distance temporal dependencies. A video clip is $m$ consecutive frames sampled by a sliding window of size $m$ in a period, and the whole video is divided into several clips. However, the content changes relatively slowly between two adjacent frames in a clip, which means the sampled consecutive frames are highly redundant. 

Similar to visual data, we observe redundancy in IMU sensor data. For instance, as shown in Fig.\ref{fig:GAF}, the IMU sensor data collected from the 3-axis accelerometer on the smartwatch has evident periodicity while the user was waving hands. We reasoned that this phenomenon should be expected in IMU sensor data. In most daily activities, human limb movements are regular and repetitive, so there is a corresponding periodicity in the data collected by wearable sensors. It means that local data features can represent the characteristics of the entire activity. Thus the complete time series is redundant. However, traditional deep learning activity recognition models based on IMU sensor data and newly proposed ones ignored this issue. In previous work, a standard input method is to feed the collected data series into deep models \cite{xu2021lIMU,wang2019deep}, such as CNN, RNN, LSTM, or Bert. Another widely-used method is dense sampling in fixed-width sliding windows and overlaps \cite{bianchi2019iot}. These unnecessary dense sampling methods lead to larger memory overhead and longer inference time. In addition, the above dense sampling strategy will consume too much energy when deployed on the device.

\citet{TSN} proposed the TSN framework to deal with frame redundancy in video understanding by applying a sparse and global temporal sampling strategy. This strategy divides the video into a fixed number of segments, and one snippet is randomly sampled from each segment. To overcome the aforementioned challenges of data redundancy in multimodal tasks, we leverage the segmentation idea of \cite{TSN} and propose a sparse sampling method for multi-wearable sensor time series, as shown in Fig.\ref{fig:sparse_sample}. 

The multi-channel grayscale images generated by GAF based on IMU sensor data have some excellent properties. First, the diagonal of each grayscale image is made of the original value of the scaled time series ($G_{i,i} = 2 \tilde{s_{t_i}}^2-1$). Second, the sampling along the diagonal direction has local temporal independence. Given two timestamps $t_i$ and $t_j$ ($i \leq j$), we sample a square area in the grayscale image with a diagonal extending from $\tilde G_{i,i}$ to $\tilde G_{j,j}$. The data in this square matrix only depends on the timestamps between $t_i$ and $t_j$, representing the original series's temporal correlation in this period. Our proposed method first divides the entire IMU sensor data into N segments of equal duration according to timestamps. The dividing points of these segments correspond to equidistant points on the diagonal of the grayscale image generated based on GAF: $\{\tilde G_{(S_0,S_0)}, \tilde G_{(S_1,S_1)}, \cdots , \tilde G_{(S_N,S_N)} \}$. In each segment, we use a square window of size $K$ for random multi-channel sampling:
\begin{equation}
\begin{aligned}
&\tilde G(S_i)=\left[\begin{array}{ccc}
\tilde G_{(i,i)} & \cdots & \tilde G_{(i,i+K-1)} \\
\vdots & \ddots & \vdots \\
\tilde G_{(i+K-1,i)} & \cdots & \tilde G_{(i+K-1,i+K-1)}
\end{array}\right],\\
\end{aligned}
\end{equation}
where ${S_{i-1}} \leq {i}$ $ \textless{i+K-1} \leq {S_i}$. We define $G(S_i)$ to be a slice of the IMU data on segment $i$. For multi-axis sensors, random segment sampling is performed simultaneously on multiple channels. 
We explore the effect of the slice length on the performance of the model (Section \ref{ablation}), and experiments show that a slice size of 2s is the optimal setting. This finding can effectively guide strategies for the sparse sampling of IMU data. 

\subsection{Inter-segment and Modality Attention Mechanism}
The attention mechanism proposed in the Transformer \cite{vaswani2017attention} has been widely applied in multimodal fusion, demonstrating significant advantages. In the context of multimodal fusion, different modalities often contain rich information, and the attention mechanism effectively enables interaction and integration of this information \cite{lu2019vilbert, islam2020hamlet}. However, the dot-product attention mechanism in the Transformer is inefficient due to its quadratic time complexity concerning the sequence length \cite{reformer, fastformer}. Fastformer \cite{fastformer} introduced additive attention, which is much more efficient than many existing Transformer models and can achieve comparable long-text modeling performance. 

\begin{figure}[h]
    \centering
    \includegraphics[width=8cm]{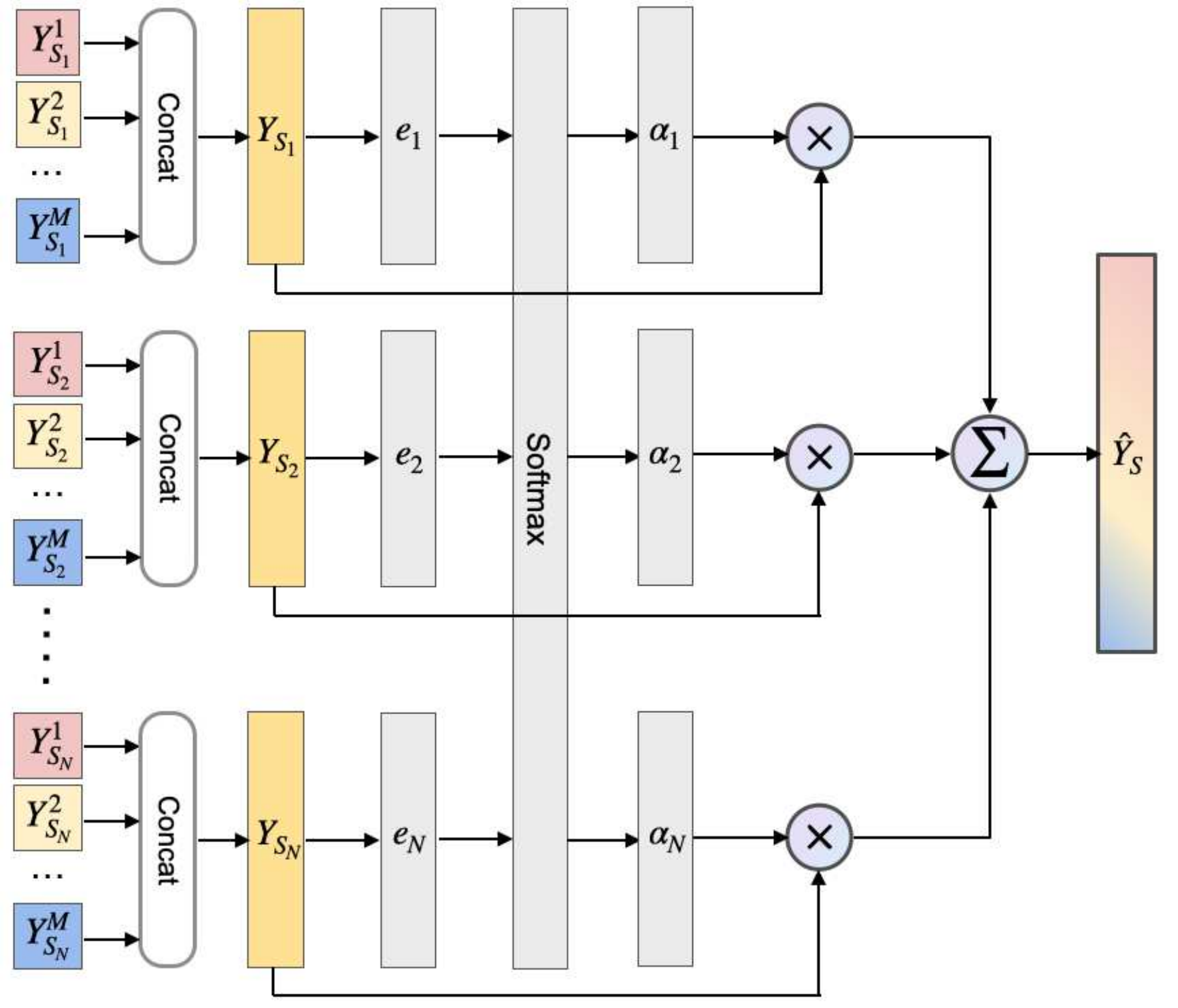}
    \caption{Inter-segment modality attention mechanism. $\mathbf{Y}_{S_i}$ is the output of the $i$-th segment. $e_i = \left(\mathbf{W}^{att}\right)^T \mathbf{Y}_{S_i}$ is a score to evaluate the importance of each segment. $\alpha_{i}$ is the regularized attention weight.} 
    \label{fig:atten}
\end{figure}

Inspired by Fastformer, we propose an efficient additive attention-based inter-segment modality fusion mechanism in MMTSA, shown in Fig. \ref{fig:atten}, to fuse features of different modal data in each segment and extract more spatiotemporal information in multimodal training. 

We first concatenate the features of different modalities in each segment by:
\begin{equation}
\mathbf{Y}_{S_i} = \mathbf{Concat}\left\{\mathbf{F_1}\left(X^1_{S_i}; \mathbf{W^1}\right), \mathbf{F_2}\left(X^2_{S_i} ; \mathbf{W^2}\right), \cdots ,\mathbf{F_m}\left(X^m_{S_i};\mathbf{W^m}\right) \right\},
\end{equation}
where $\mathbf{Y}_{S_i}$ is the output of the $i$-th segment, $\mathbf{F_j}\left(X^j_{S_i} ; \mathbf{W^j}\right)$ represents a ConvNet with parameters $\mathbf{W^j}$ that operates on $X^j_{S_i}$ and $j$ indicates the $j$-th modality. $X^j_{S_i}$ and $\mathbf{W^j}$ represent the sampled data of the $j$-th modality in segment $i$ and the shared parameters of modality $j$, respectively. Then we get an output sequence of each segments: $\mathbf{Y}^{out}=\left(\mathbf{Y}_{S_1}, \mathbf{Y}_{S_2}, \cdots, \mathbf{Y}_{S_N} \right)$ where 
$\mathbf{Y}_{S_i} \in \mathbb{R}^{(m \times d)}$ 
and $d$ is the feature dimension of each modality.

We utilize additive attention to calculate the attention weight of each segment,
\begin{equation}
\mathbf{\alpha}_{i} =\frac{\exp \left(\left(\mathbf{W}^{att}\right)^T \mathbf{Y}_{S_i}\right)}{\sum_{i \in N} \exp \left(\left(\mathbf{W}^{att}\right)^T \mathbf{Y}_{S_i}\right)},
\end{equation}
where $\mathbf{W}^{att} \in \mathbb{R}^{(m \times d)}$ is a learnable parameter and $\left(\mathbf{W}^{att}\right)^T \mathbf{Y}_{S_i}$ is a score to evaluate the importance of each segment. The softmax function is used to regularize the scores so that the sum of the scores of all segments is 1. Here, the parameter $\mathbf{W}^{att}$ will weigh the features of different modalities to compensate for the inter-modal information.

Next, the attention weights $\alpha=\{\alpha_{1},\alpha_{2},\cdots,\alpha_{N}\}$ will be used to fuse the outputs of each segment and get a weighted global representation,
\begin{equation}
\tilde {\mathbf{Y}_{S}}=\sum_{i \in N} \alpha_{i} \mathbf{Y}_{S_i}.
\end{equation}

Finally, the global representation $\tilde {\mathbf{Y}_{S_i}}$ is fed into a Feed Forward Neural Network with two fully-connected layers and a softmax to compute the probabilities for each class of human activities.

\section{Experiment}
\subsection{Dataset}
We evaluate our proposed model multimodal temporal segment attention network MMTSA for the human activity recognition task on three public datasets: MMAct \cite{Kong_2019_ICCV}, DataEgo \cite{possas2018egocentric}, and Multimodal Egocentric Activity \cite{song2014activity}.

\textbf{MMAct \cite{Kong_2019_ICCV}}: Compared with the other two datasets, the dataset is a large-scale multimodal action dataset consisting of more than $36000$ trimmed clips with seven modalities performed by 20 different subjects, which include RGB videos, acceleration, gyroscope, orientation, keypoints, WiFi, and pressure signal. Each modality has 37 action classes. Moreover, for each activity, it provides five camera views in total. Four were recorded from 4 top corners of the space, and one was recorded from the egocentric view by wearing the smart glass. In this paper, we use RGB videos (30 fps) and four different wearable-sensors modalities, accelerator-phone (100Hz), acclerator-watch (100Hz), gyroscope (50Hz), and orientation (50Hz). We use two different settings to evaluate this dataset: cross-subject, and cross-session, according to the train-test split strategy described in the original MMAct dataset paper.

\textbf{DataEgo \cite{possas2018egocentric}}: The dataset contains 20 distinct activities performed in different conditions and by 42 different subjects. Each recording has 5 minutes of footage and contains a sequence of 4-6 different activities to enable a natural flow between different activities. Moreover, the data is captured with the Vuzix M300 Smart Glasses. Images from the camera are synchronized with readings from the accelerometer and gyroscope captured at 15 fps and 15 Hz, respectively. DataEgo is an egocentric dataset composed of three modalities (RGB videos, gyroscope, and accelerometer) and contains approximately 4 hours of continuous activity.

\textbf{Multimodal Egocentric Activity \cite{song2014activity}}: The dataset contains 20 different life-logging activities performed by different human subjects. Each activity category has ten sequences. Each clip has precisely 15 seconds. Moreover, the egocentric videos (29.9 fps) are augmented with rich sensor signals (10 Hz), which include an accelerometer, gravity, gyroscope, linear acceleration, magnetic field, and rotation vector. 

\subsection{Experiment Settings}
We compare our proposed model multimodal temporal segment attention network with the following state-of-the-art multimodal HAR training algorithms for comparison, such as TSN \cite{TSN}, Keyless \cite{long2018multimodal}, HAMLET \cite{islam2020hamlet}, MuMu \cite{islam2022mumu}. We use the micro F1 score to evaluate the performance of all methods. 


\subsubsection{Video Data Processing}: In our implementation, one input video is divided into $N$ segments equally, and an RGB frame is randomly selected from each segment. 
The frames are rescaled and center-cropped to a size of 224*224, which fits the CNN input requirements. The sparse-sampled RGB frames from $N$ segments represent the input of visual modality.

\subsubsection{IMU Sensor Data Processing}: It is redundant and has high memory costs to encode the whole IMU sensor data into the GAF-based grayscale image and sample several square pixel matrix regions along the diagonal of it. Alternatively, to achieve the efficiency of our method, we directly segment the IMU time series equally and select a data sequence with a duration of K timestamp within each segment randomly. The size of K depends on the sampling rate of the sensor, and K multiplied by the sampling rate is equal to a uniform predefined slice length. Then we convert each sequence to a grayscale image based on GAF. To keep synchronization, the IMU data series is divided into $N$ segments, the same as the number of video data segments. Therefore, it is efficient and equivalent to encode the IMU wearable sensor data into the grayscale image sparsely, as in Sec.\ref{sbmss}. Furthermore, since most wearable sensors are triaxial, by combining the grayscale image from each axis, the chosen multi-channel grayscale images from $N$ segments represent the IMU wearable sensor data input modality.

\subsubsection{Training Details:} We implement our proposed method in Pytorch 1.7. We utilize Inception with Batch Normalization (BN-Inception) as a sub-CNN to extract the unimodal feature representations. Moreover, in the proposed model, we trained all the modalities simultaneously with $N=3$ segments, SGD with momentum optimizer, and a learning rate of $0.001$. The convolutional weights for each modality are shared over the $N$ segments, reducing the model size and memory cost.

\section{Result and Discussion}
\subsection{Results Comparison} \label{sec:result-comp}

We evaluate our proposed model MMTSA performance and summarize all the results. For the MMAct dataset, we follow the proposed initial cross-subject and cross-session evaluation settings and report the results in Table \ref{mmact}. The results show that MMTSA improves $11.13\%$ and $2.59\%$ in cross-subject and cross-session evaluation settings \cite{Kong_2019_ICCV}. The other methods compared in the table also follow the division criteria of the training set and test set in the original paper \cite{Kong_2019_ICCV} of the MMAct dataset. The experimental results of these methods refer to Mumu \cite{islam2022mumu} and Multi-GAT \cite{islam2021multi}. For DataEgo data, we divide each 5-minute original data into 15 seconds clips with 5-second overlapping. We keep the train and test split size of each activity balance. The performance of our method is shown in Table \ref{datageo}, which outperforms TSN by $17.45\%$. Given that the source code and some implementation details for Hamlet \cite{islam2020hamlet}, Mumu \cite{islam2022mumu}, and Multi-GAT \cite{islam2021multi} are absent, we reproduced these three models, adjusted their parameters, and trained them from scratch using the same training and testing dataset splits. For Multimodal Egocentric Activity, we follow leave-one-subject-out cross-validation. MMTSA outperforms all of the traditional methods and is close to the performance of MFV, as shown in Table \ref{egocentric}. Compared to MFV, which uses four types of sensor data, MMTSA only uses two types (accelerometer, gyroscope) as input to make the model lightweight. Thus, a small loss of precision is acceptable. 
\begin{table}[!htp]
\centering
\caption{cross-subject and cross-session performance comparison on MMAct dataset}
\label{table:1}
\begin{tabular}{cccc}
\hline
Cross-Suject  & F1-Score (\%) & Cross-Session  & F1-Score (\%)  \\ \hline
Multi-Teachers \cite{Kong_2019_ICCV}                         & 63.89                       & SVM+HOG \cite{ofli2013berkeley}                     & 46.52                       \\
Student \cite{Kong_2019_ICCV}              & 62.27                       & TSN (Fusion) \cite{TSN}                & 77.09                       \\
MMAD \cite{Kong_2019_ICCV}                     & 64.44                       & MMAD \cite{Kong_2019_ICCV}                        & 78.82                       \\
HAMLET \cite{islam2020hamlet}                       & 69.35                       & Keyless \cite{long2018multimodal}                     & 81.11                       \\
Keyless \cite{long2018multimodal}                      & 71.83                       & HAMLET \cite{islam2020hamlet}                      & 83.89                       \\
Multi-GAT \cite{islam2021multi}                    & 75.24                       & MuMu \cite{islam2022mumu}                        & 87.50                       \\
MuMu \cite{islam2022mumu}                        & 76.28                       & Multi-GAT \cite{islam2021multi}                    & 91.48                       \\
\textbf{MMTSA (our method)} & \textbf{87.41}              & \textbf{MMTSA (our method)} & \textbf{94.07}                       \\ \hline
\end{tabular}%
\label{mmact}
\end{table}

We notice that models, such as Mumu, and Multi-GAT, that perform well on the MMAct dataset have relatively poor performances on the DataEgo dataset. We attribute this to differences in the data modality and subjects' environment of the two datasets. The video data in MMAct is collected from four RGB cameras fixed in the upper corners of the same room, and the video data are all from the third-person perspective. The data collected in this way lacks the interference of environmental information and camera shake, and the human body's complete posture and body movements can be clearly captured. However, the video data in DataEgo is all first-person perspective, and the camera that records the data is located on the AR glasses, which brings two challenges. First of all, the complete body posture of the subject is missing in the video, and often only the hand movements are captured. Secondly, camera shake and environmental interference are apparent. For example, the information recorded while running and walking is the outdoor environment seen by the subjects. For IMU data, DataEgo's sensors are in AR glasses, while MMAct's sensors are in smartwatches and smartphones. Since in most daily activities of humans, the movement of hands and legs is more abundant and violent than that of the head, the effective information of IMU data in DataEgo is relatively limited. The results in Table \ref{datageo} show that the Mumu and Multi-GAT cannot well recognize human activities for first-person perspective data has much redundant information and lacks human pose information. The TSN model performs well due to the sparse sampling strategy for video data, which avoids the input of redundant information and a large amount of noise. The MMTSA we proposed has better recognition results in both the DataEgo dataset and the MMAct dataset than other methods, which shows that MMTSA has stronger robustness and generalization capabilities and can effectively model multimodal data from different devices.

\begin{table}[!htp]
\centering
\caption{Cross-subject performance comparison on Multimodal Egocentric Activity}
\label{table:1}
\begin{tabular}{cc}
\hline
Method                    & F1-Score (\%)  \\ \hline
SVM \cite{kwapisz2011activity}                      & 47.75          \\
Decision Tree \cite{kwapisz2011activity}                      & 51.80          \\ 
FVS \cite{song2016egocentric}                      & 65.60          \\    
TFVS \cite{song2016egocentric}                      & 69.00          \\
Multi-Stream with average pooling \cite{song2016multimodal}                      & 76.50          \\
FVV \cite{song2016egocentric}                      & 78.44          \\
FVV+FVS \cite{song2016egocentric}                      & 80.45          \\
Multi-Stream with maxIMUm pooling \cite{song2016multimodal}                  & 80.50          \\
\textbf{MMTSA (our method)} & \textbf{80.50} \\
MFV \cite{song2016egocentric}                  & 83.71          \\ \hline

\end{tabular}
\label{egocentric}
\end{table}

\begin{table}[ht]
\centering
\caption{Cross-subject performance comparison on DataEgo}
\begin{tabular}{cc}
\hline
Method                    & F1-Score (\%)  \\ \hline
Multi-GAT \cite{islam2021multi}                    & 19.69 \\
HAMLET \cite{islam2020hamlet}                      & 21.03 \\
MuMu \cite{islam2022mumu}                        & 22.15 \\
TSN (RGB) \cite{TSN}                      & 65.77          \\

\textbf{MMTSA (our method)} & \textbf{83.22} \\ \hline
\end{tabular}
\label{datageo}
\end{table}

\subsection{Ablation Study} \label{ablation}
This section is organized as follows. First, we demonstrate the necessity of multimodal data fusion by comparing the impact of different inputs of single modality and multimodality on the performance of human activity recognition. Second, we compare the effects of direct feature concatenation and additive inter-segment attention in the feature fusion stage. In this section, all experiments are done on the MMact dataset and the DataEgo dataset. The training-test set split of the MMAct dataset follows the cross-subject approach.

\subsubsection{Effectiveness of Multimodal Fusion} To investigate the importance of the multimodal fusion of MMTSA, We compare the performance on the MMAct dataset when taking a single modality or a combination of multiple modalities as input, see Figure \ref{modal-com}. The modalities we evaluate include RGB video, accelerometer data of smartphone (denoted as A1), accelerometer data of smartwatch (denoted as A2), gyroscope data (denoted as G), and orientation data (denoted as O). Combining inputs of different types of modalities yields better recognition results than using a single visual modality or IMU-based modality as input. It indicates that the multimodal isomorphism and fusion mechanism in MMTSA mines the complementary information among modalities more comprehensively. 

\begin{figure}[!htbp]
	\centering
	\subfigure[Results of input modalities]{
		\begin{minipage}[b]{0.48\linewidth}
            \centering
			\includegraphics[width=1\linewidth]{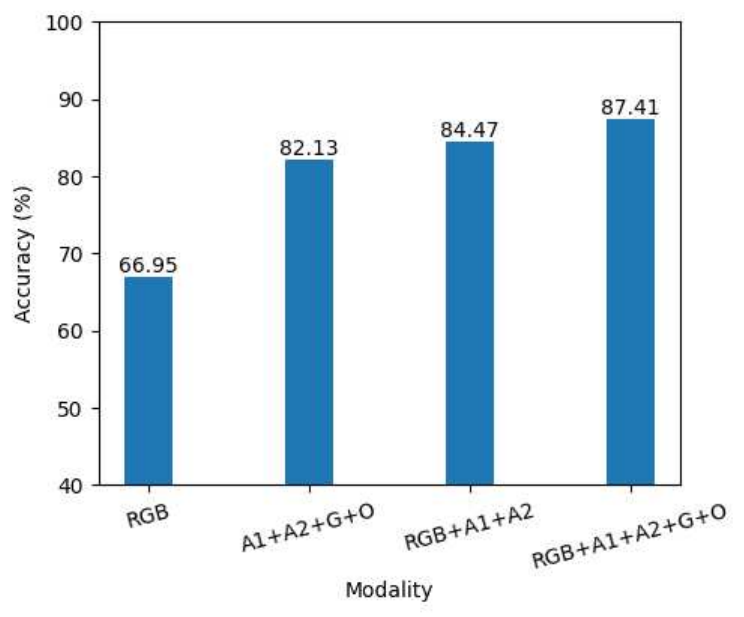} 
		\end{minipage}
		\label{modal-com}
  }
    \subfigure[Results of fusion methods]{
        \begin{minipage}[b]{0.3\textwidth}
            \centering
            \includegraphics[width=1\textwidth]{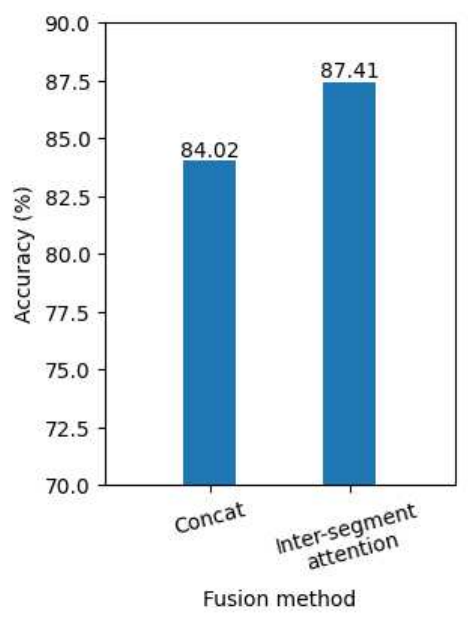}
        \end{minipage}
        \label{att-com}
        }
	\caption{Ablation study on MMAct (cross-subject)}
	\label{fig:acclero}
\end{figure}

\begin{figure}[h]
	\centering
	\subfigure[RGB video features]{\label{fig:vis-r}\includegraphics[width=0.33\linewidth]{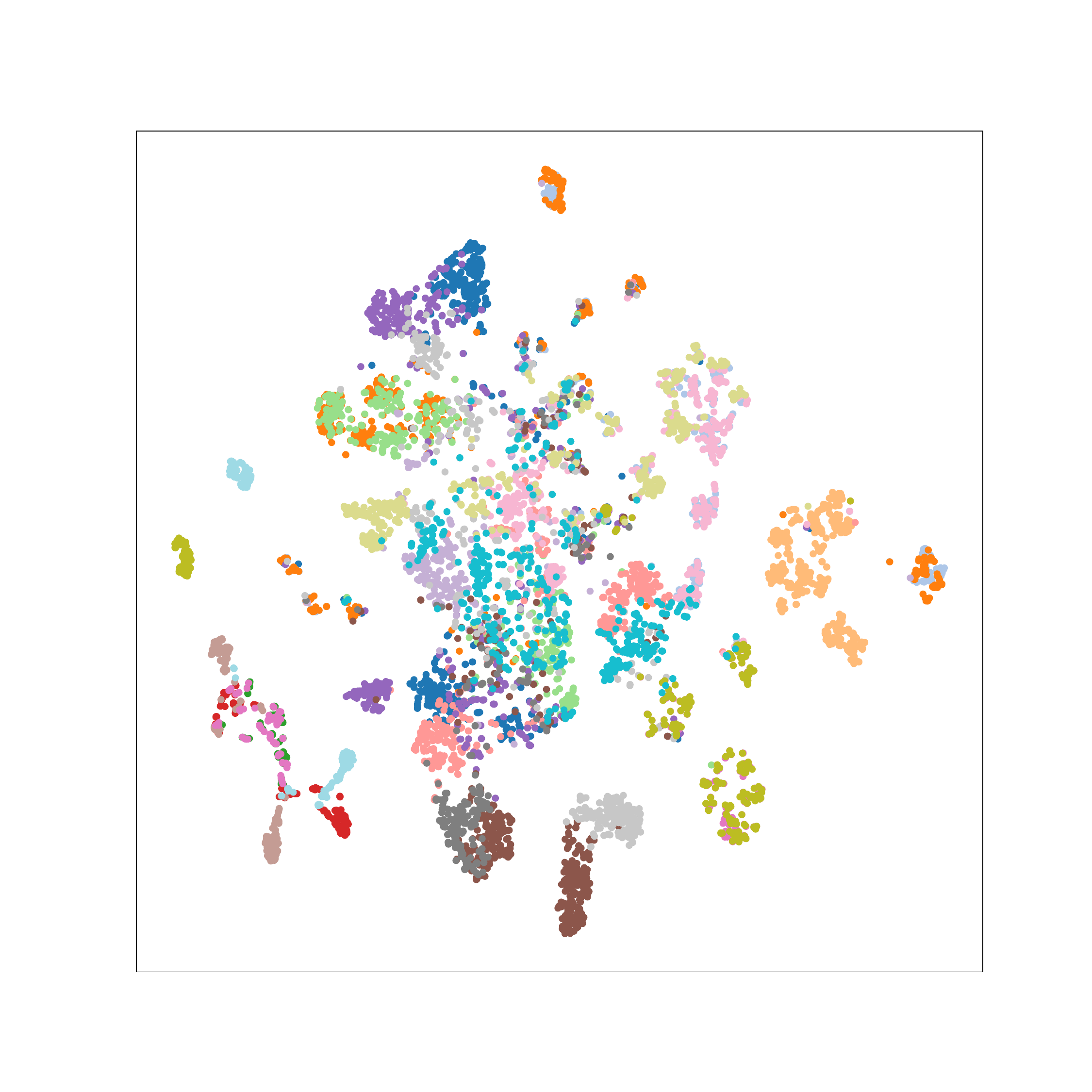}}
	\subfigure[IMU features]{\label{fig:vis-s}\includegraphics[width=0.33\linewidth]{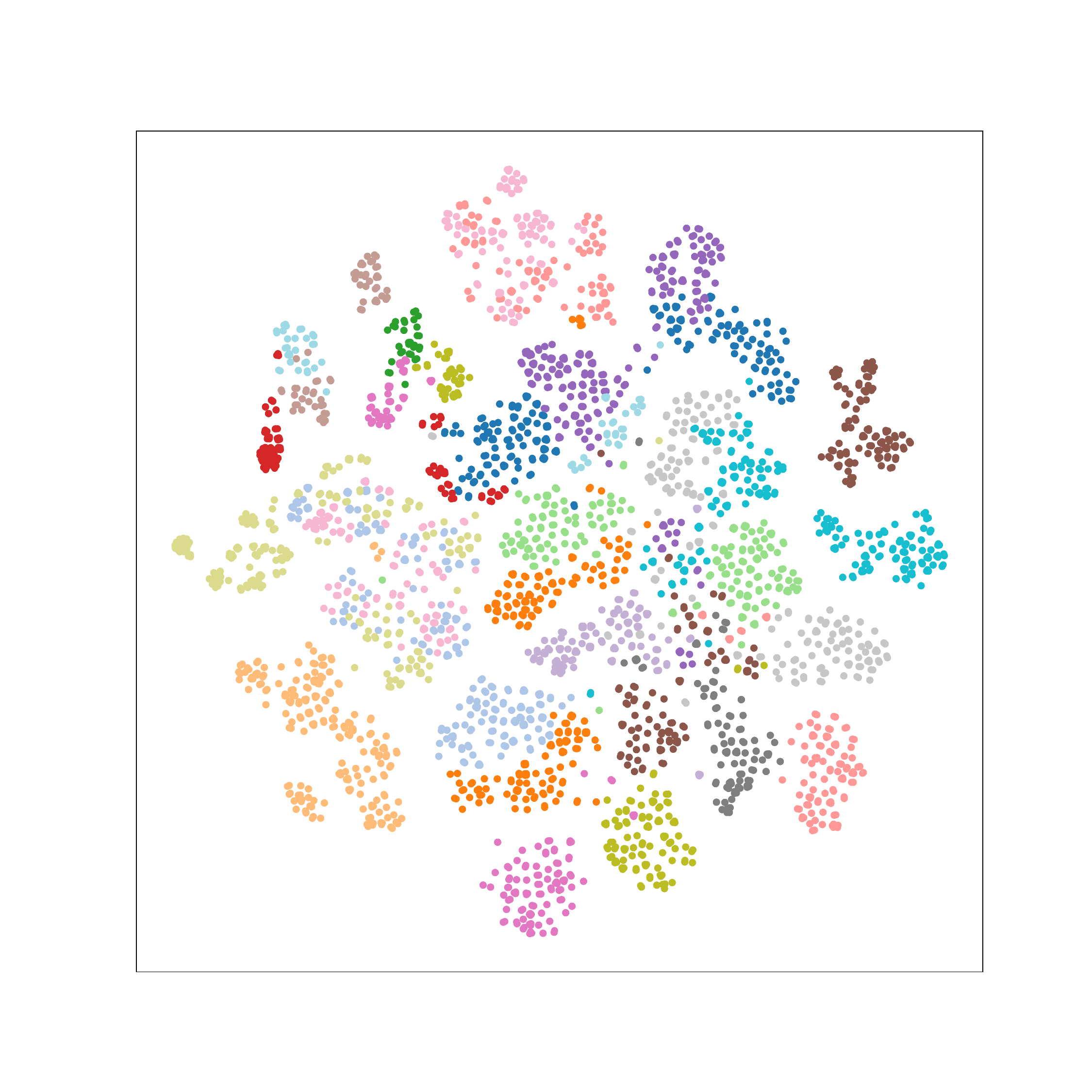}}
	\subfigure[MMTSA Multimodal features]{\label{fig:vis-rs}\includegraphics[width=0.33\linewidth]{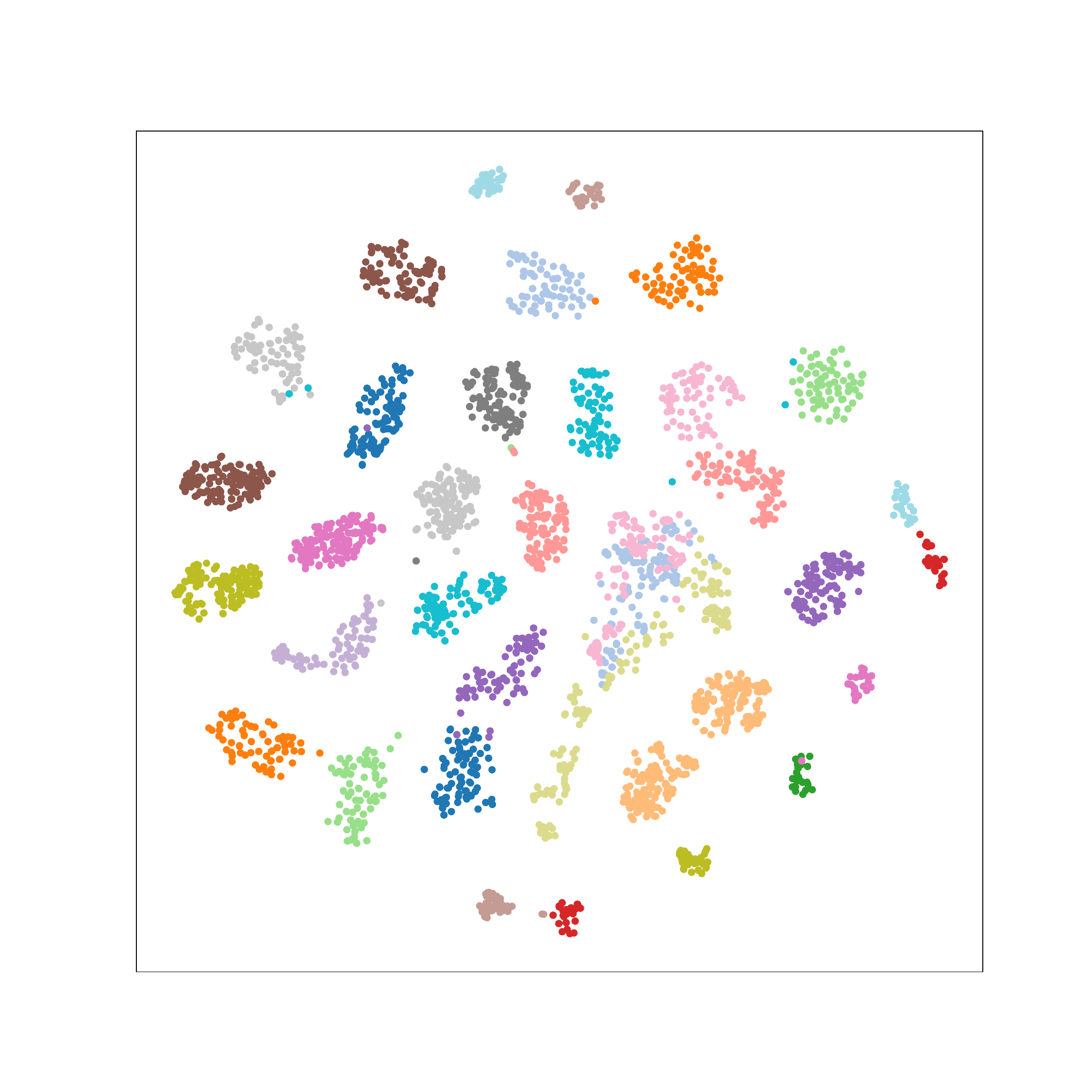}}
	\caption{2-D t-SNE embeddings of features for the MMAct dataset. A single marker represents a single activity clip and is color-coded by its type. (a) Embeddings of RGB videos features. (b) Embeddings of IMU features. (c) Embeddings of multimodal features fusion. All the features are extracted from the last FC layer of MMTSA.}
	\label{fig:vis-mmact}
\end{figure}

We also qualitatively evaluate our proposed model by visualizing various modality embeddings for the MMAct dataset. We input different modal data and extract output features of the last fully connected layer from the best-performing MMTSA, and project them to 2-dimensional space using t-SNE. Figure \ref{fig:vis-mmact} clearly shows that multimodal features extracted from the last FC layer are more discriminative than either RGB video features or IMU features. This further indicates that MMTSA is capable of fusing consistent and complementary information from different modalities to enable effective feature extraction.

\subsubsection{Effectiveness of Inter-segment Modality Attention Mechanism} We compare the simple concatenation method and our proposed inter-segment attention fusion method in MMTSA. The input modalities of the two experiments are RGB+A1+A2+G+O. The results in \ref{att-com} suggest that additive inter-segment attention outperforms simple concatenation. It indicates that the inter-segment attention modality fusion method helps MMTSA to more effectively extract the consistency and complementarity information between modalities. In addition, we also find that after using inter-segment attention, the model converges faster during training.

\subsubsection{Effectiveness of GAF-based Imaging Method} 
\begin{figure}[h]
	\centering
	\subfigure[Performance of using different encoders for IMU data]{\label{fig:result-trans}\includegraphics[width=0.32\linewidth]{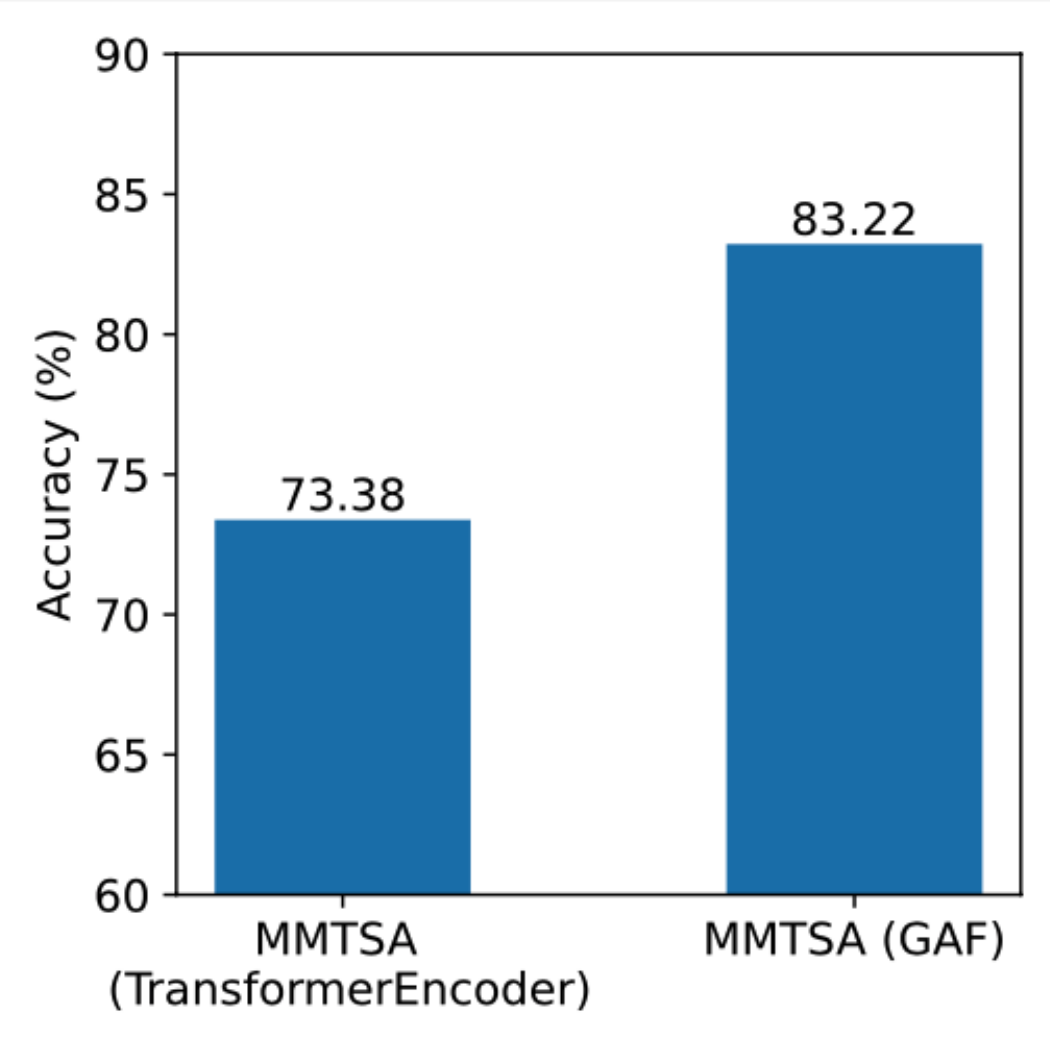}}
	\subfigure[2-D t-SNE embeddings of GAF-based encoder]{\label{fig:tsne-g}\includegraphics[width=0.31\linewidth]{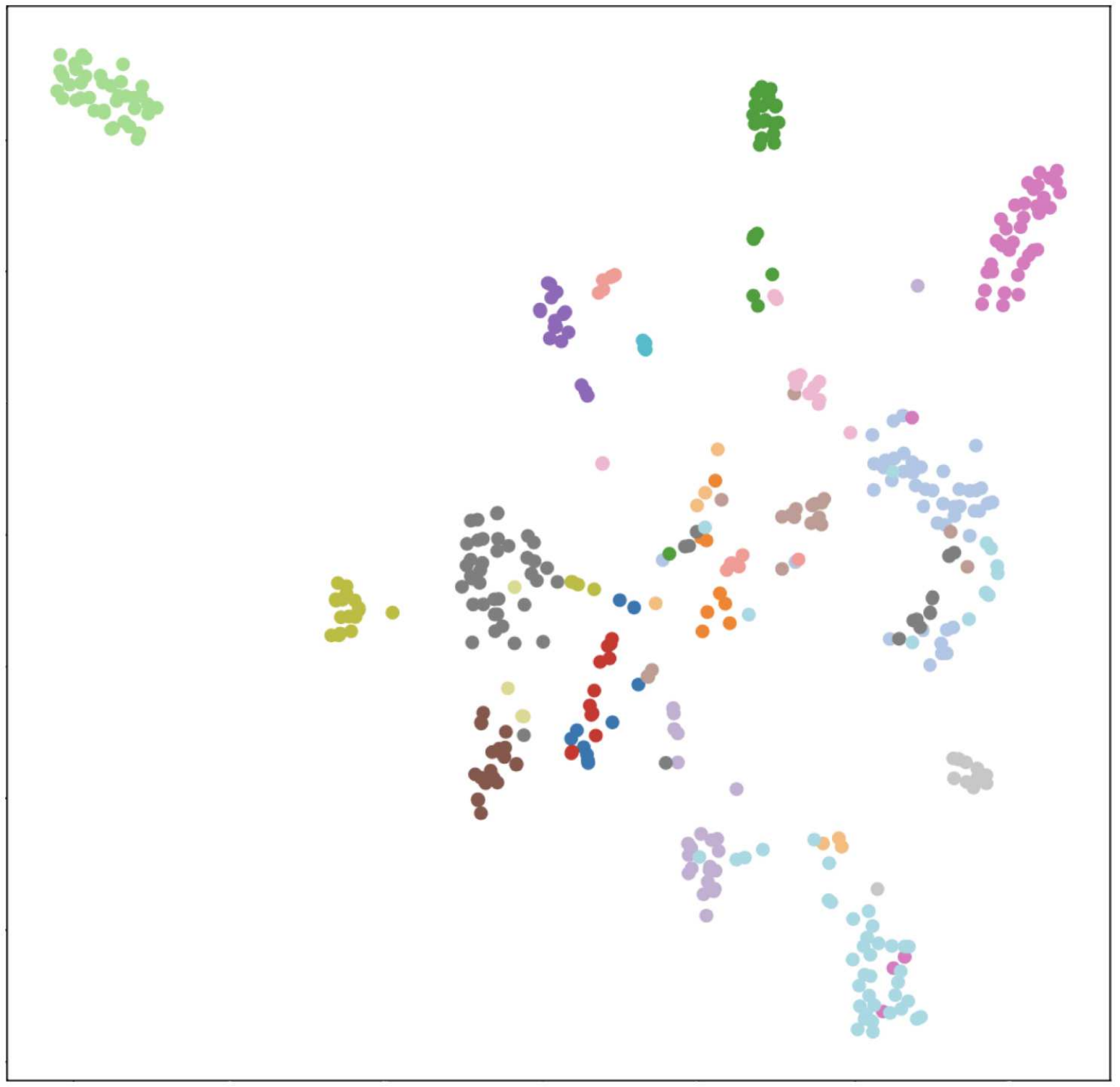}}
	\subfigure[2-D t-SNE embeddings of transformer-based encoder]{\label{fig:tsne-t}\includegraphics[width=0.3\linewidth]{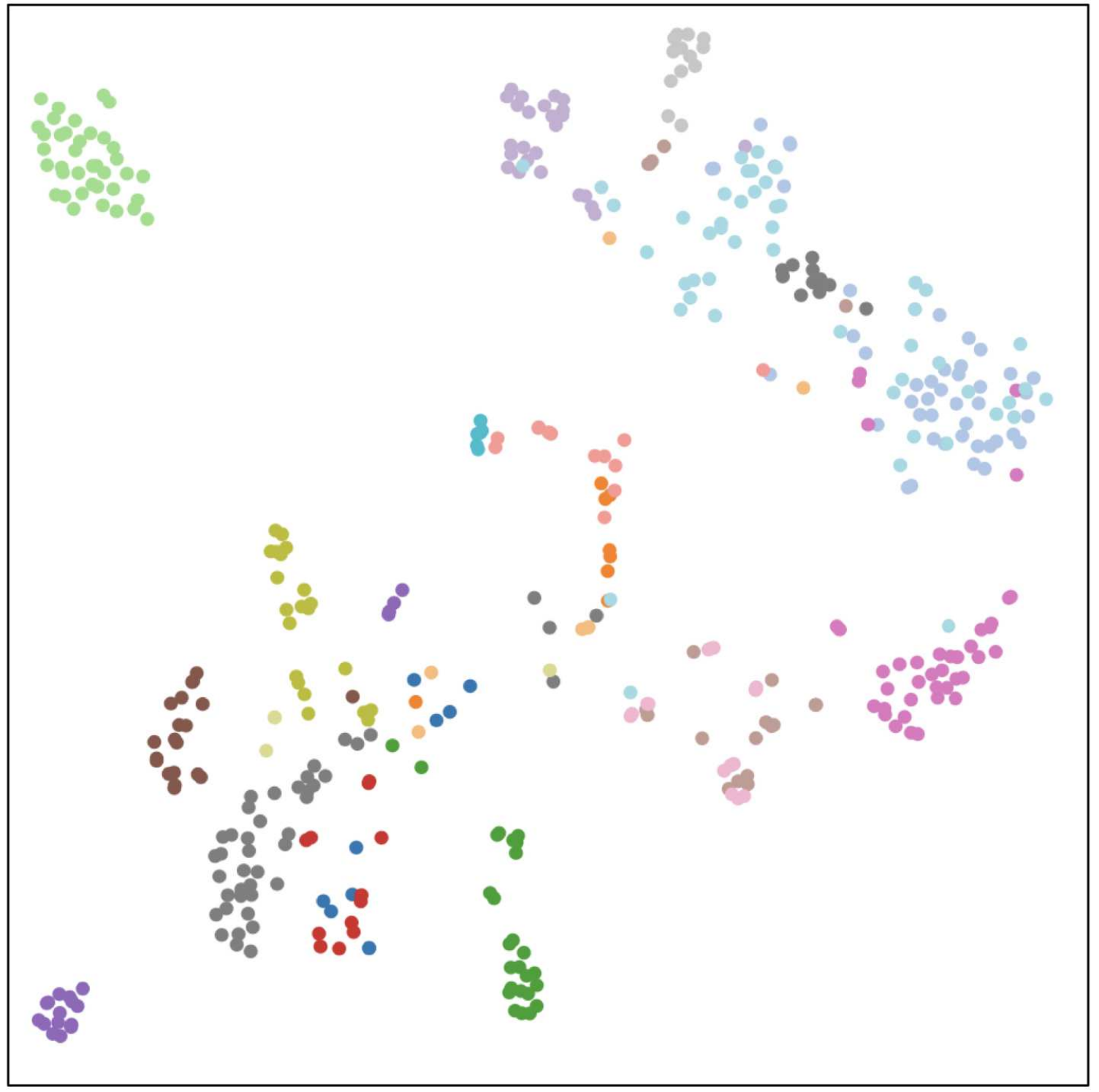}}
	\caption{Ablation study on DataEgo.}
	\label{fig:vis-DataEgo}
\end{figure}

To verify the effect of the GAF-based imaging method in our model, we perform experiments with different encoders for IMU data on the Dataego dataset. The transformer architecture has been extensively employed as an IMU data encoder in state-of-the-art human daily activity recognition algorithms. Hence, we opt for a representative transformer encoder architecture\cite{shavit2021boosting} to replace the GAF-based IMU encoding module in MMTSA and compare the recognition performance with the original MMTSA. The results in Figure \ref{fig:result-trans} show that the GAF-based imaging method outperforms the transformer-based encoding in our proposed architecture. Figure \ref{fig:tsne-g} and Figure \ref{fig:tsne-t} show the t-SNE visualization results of the embedding representation obtained through the GAF-based imaging method and a transformer encoding method, respectively. Compared with the features extracted by the transformer encoder, the features extracted based on the GAF imaging method are visually more separable in different activities, and the data points of the same activity are more clustered. These findings indicate that the GAF-based imaging method can more effectively model the IMU data and extract more discriminative features. We believe that the GAF-based method enhances certain waveform patterns and physical semantics in IMU data compared to Transformer-based encoders, and these features are more easily captured by 2D CNNs. Transformer-based encoders are more suitable for scenarios with longer IMU sequences and enough training data.

\subsubsection{Length of IMU Data Slice for Multimodal Sparse Sampling}
To understand the importance of the sparse sampling strategy on multimodal data, we conduct experiments on the DataEgo and MMAct datasets with different lengths of IMU sampling slices. Table \ref{abl-len} shows that sampling an IMU slice of 2s can achieve the best performance. We can conclude that longer sampling slices will introduce unnecessary redundant information, while shorter ones may miss enough valuable features. Choosing a sampling IMU slice of 2 seconds is an ideal sparse sampling strategy for the HAR task.
\begin{table}[h]
\caption{Ablation study on the length of IMU slice on MMAct and DataEgo dataset.}
\begin{tabular}{ccccccc}
\hline
\multirow{2}{*}{} & \multicolumn{3}{c}{\begin{tabular}[c]{@{}c@{}}MMAct \\ (Cross-subject)\end{tabular}} & \multicolumn{3}{c}{DataEgo}                                             \\ \cline{2-7} 
                  & \multicolumn{1}{c}{T@1}      & \multicolumn{1}{c}{T@2}                & T@3     & \multicolumn{1}{c}{T@1} & \multicolumn{1}{c}{T@2}            & T@3 \\ \hline
F1-Score (\%)     & \multicolumn{1}{c}{83.47}      & \multicolumn{1}{c}{\textbf{87.41}}     & \multicolumn{1}{c}{84.84}          & \multicolumn{1}{c}{68.23} & \multicolumn{1}{c}{\textbf{83.22}} & 79.42 \\ \hline
\end{tabular}
\label{abl-len}
\end{table}

\subsection{Analysis of Extracted Information}
To understand the effectiveness of MMTSA and its multimodal fusion method, we show the recognition accuracy of MMTSA for different activities in the DataEgo as well as the confusion matrix when single modality data is input.

\begin{figure}[h]
	\centering
	\subfigure[Confusion matrix of visual modality input]{
		\begin{minipage}[b]{0.48\linewidth}
            \centering
			\includegraphics[width=1.2\linewidth]{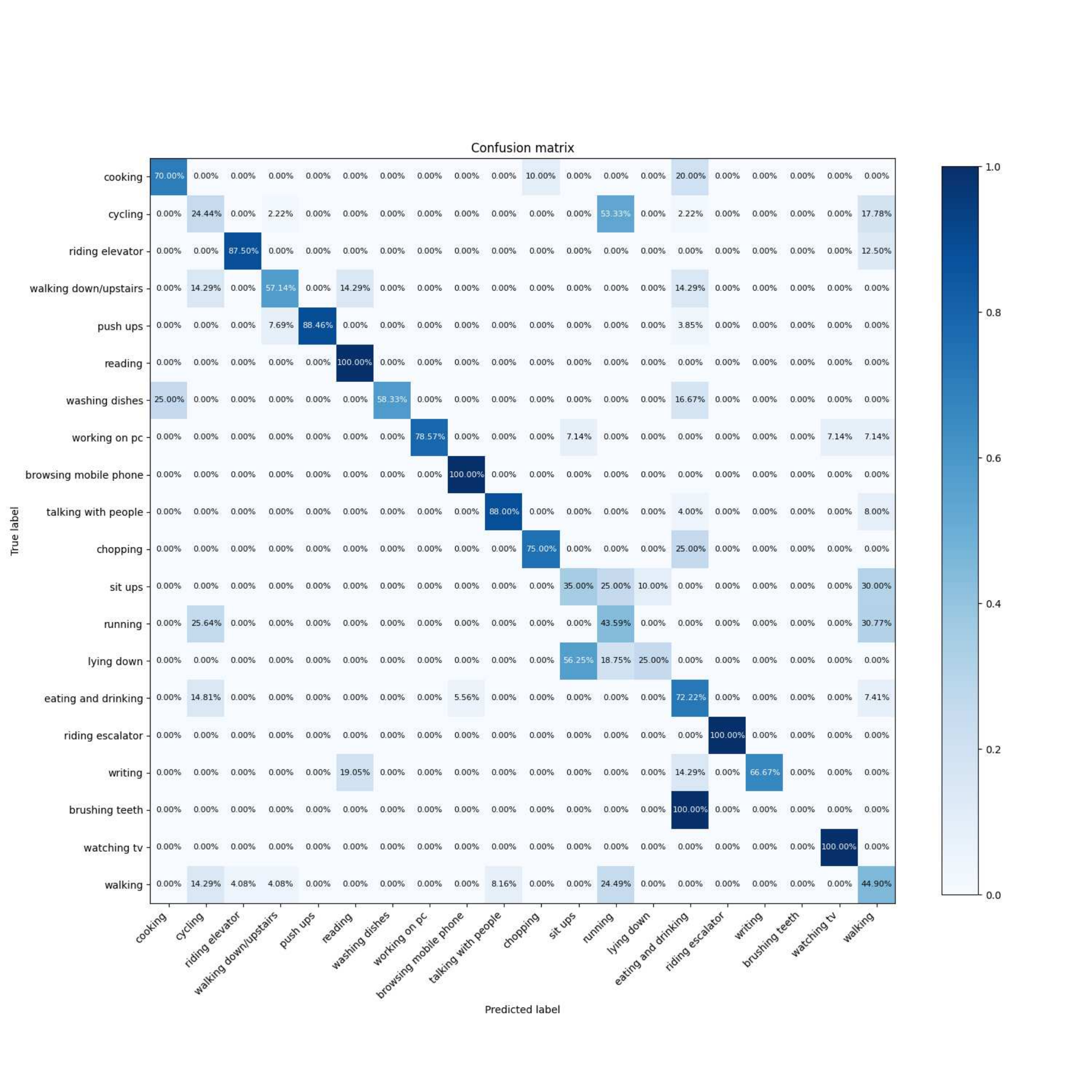} 
		\end{minipage}
		\label{modal-com}
  }
    \subfigure[Confusion matrix of IMU-based modality input]{
        \begin{minipage}[b]{0.48\linewidth}
            \centering
            \includegraphics[width=1.2\linewidth]{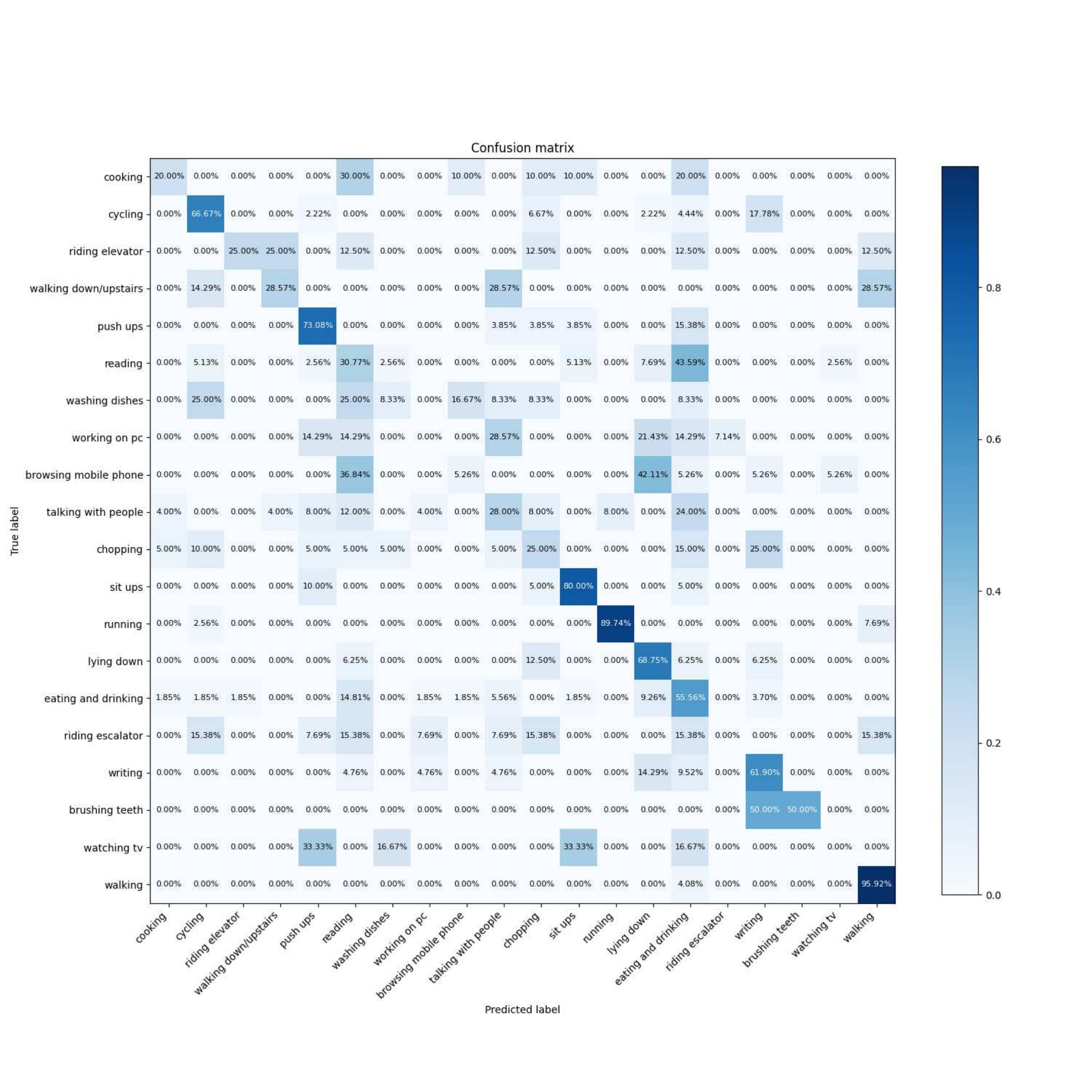}
        \end{minipage}
        \label{att-com}
        }
    \\
    \centering
	\subfigure[MMTSA extracts the complementarity information of multimodal data]{
		\begin{minipage}[b]{0.8\linewidth}
            \centering
			\includegraphics[width=1.0\linewidth]{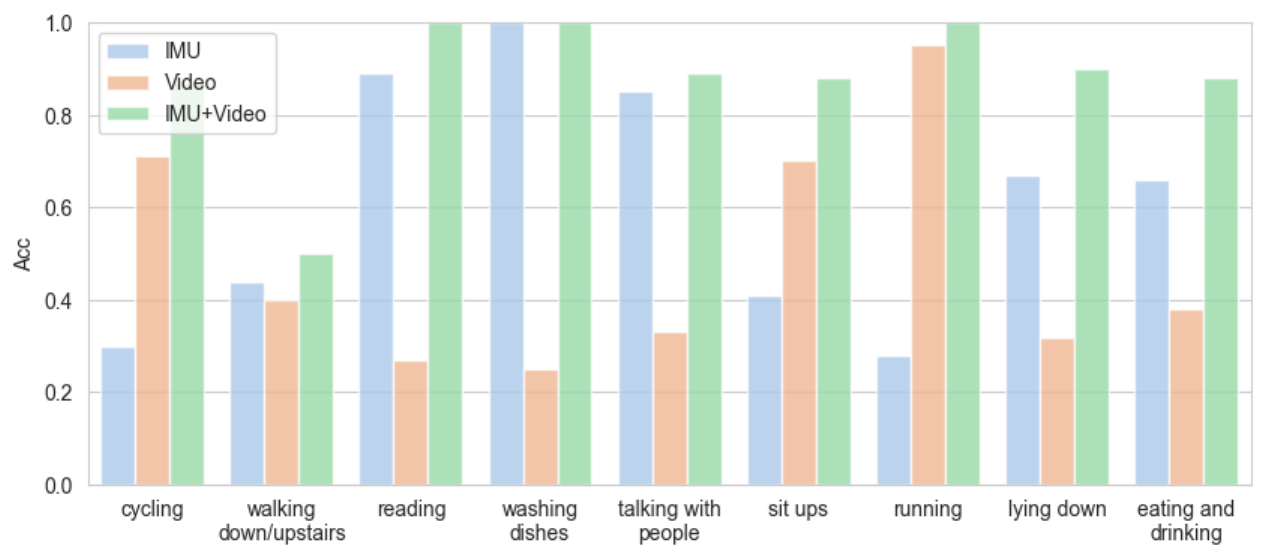} 
		\end{minipage}
		\label{bar}
  }
    \caption{Recognition performance of MMTSA on each activity in Dataego}
\end{figure}

Figure \ref{modal-com} and \ref{att-com}, respectively, show the confusion matrix of the MMTSA recognition results when the visual modality and the IMU-based modality are used as input alone. We find the complementarity of these two different modality data. Visual modality performs better on some relatively static activities, such as watching TV, browsing mobile phones, working on PC, talking with people, etc. The position of subjects' heads is relatively fixed in such activities, so the practical information contained in the sensor data is relatively scarce. However, the recognition performance of IMU-based modality is significantly better in some dynamic activities. For instance, when using visual information alone to identify cycling and walking, the model tends to confuse them with running. In contrast, the IMU-based modality helps the model clearly identify these three activities. We believe that when the above three activities were performed, the video data recorded by the AR glasses shook obviously and the surrounding environment was similar, so the visual data lacks the specific information of these three activities. IMU-based modality addresses this challenge well. Likewise, visual data can easily confuse lying down and doing sit-ups, whose practical motion information is included in the IMU-based modality, due to the similarity of viewing angles. Figure \ref{bar} compared the recognition accuracy of MMTSA on several activities when taking different modalities as input. We select several activities in which the performance gap between the two modal inputs is large or both are poor. When both modalities are used as input to MMTSA, the recognition results of these activities are better than when a single modality is an input. It shows that MMTSA can comprehensively extract consistent information and complementary information between modalities and that the multimodal fusion mechanism of MMTSA is effective.

\subsection{FLOPs and Latency of Edge Deployment}


\begin{table}[!htbp]
\caption{Evaluation results of model performance and efficiency}
\begin{tabular}{cccclcc}
\hline
\multirow{2}{*}{Model}     & \multicolumn{3}{c}{Efficiency (DataEgo)}                     &  & \multicolumn{2}{c}{Accuracy (MMTSA)}                         \\ \cline{2-4} \cline{6-7} 
                           & FLOPs            & Param.          & Latency       &  & Cross-Session    & \multicolumn{1}{l}{Cross-Subject} \\ \hline
HAMLET \cite{islam2020hamlet}                     & 311.158G         & 25.41M          & 34.7s         &  & 83.89\%          & 69.35\%                                \\
MuMu \cite{islam2022mumu}                      & 311.036G         & 25.19M          & 33.5s         &  & 87.50\%          & 76.28\%                           \\
Multi-GAT \cite{islam2021multi}                 & 313.015G         & 32.00M          & 35.9s         &  & 91.48\%          & 75.24\%                           \\
\textbf{MMTSA (our model)} & \textbf{18.428G} & \textbf{32.39M} & \textbf{5.8s} &  & \textbf{94.07\%} & \textbf{87.41\%}                  \\ \hline
\end{tabular}%
\label{effe-tab}
\end{table}

To evaluate and compare MMTSA's performance and efficiency to the state-of-art methods, we deployed the model on an embedded system --- a single Raspberry Pi 4B with a 64-bit 1.5GHz 4-core CPU (ARM Cortex-A72) and 8-GB RAM. 
We measured the latency, FLOPs, and parameter amount as the metrics on the DataEgo dataset. We selected DataEgo because its preprocessed clips have consistent durations, ensuring experiment fairness.
The input modalities we used are RGB videos, smartphone accelerometer data, and smartwatch accelerometer data. Each input last for 15 seconds.
The batch size is set to 1. 
As Table \ref{effe-tab} shows, although MMTSA possesses slightly more parameters, it has significantly lower FLOPs and latency but higher accuracy than previous state-of-the-art models, including HAMLET, MuMu, and Multi-GAT. In conclusion, only requiring 6\% of FLOP, MMTSA achieves higher recognition accuracy and reduces latency by  82.6\%  when compared with state-of-the-art models. 
Thus, MMTSA is more friendly to edge deployment.

\section{Limitation and Future Work}
In this section, we discuss the limitations of this paper and outlook future research directions.
\paragraph{Extensions to New Modalities:}
In this paper, we only focus on designing MMTSA to HAR utilizing IMU and RGB data. However, it should be noted that MMTSA exhibits scalability towards other modalities as well. The GAF-based imaging method is well-suited for a wide range of one-dimensional modalities, such as heart rate, photoplethysmography signals, light intensity, sound, and so on. By employing the isomorphism method described in Section \ref{gaf_image}, these types of data can all be encoded into two-dimensional grayscale images. To enhance the interpretability of this extended approach, future research should also address the correspondence between the generated GAF images and the physical meanings of other modal data.

\paragraph{Automated Segmentation and Feature Selection:} Another limitation exists in MMTSA's sparse sampling strategy. In MMTSA, the number of segments of input clips and the length of each IMU slice are predetermined as hyperparameters. We investigate the optimal sampling strategy through ablation experiments, and MMTSA demonstrates superior performance compared to other SOTAs in HAR when employing this strategy. Nevertheless, the hand-generated sampling configuration may be too rigid. In the real world, the model structure and configuration should be adjusted to match the specific activities. Therefore, it is necessary for MMTSA to introduce an automatic configuration mechanism and feature selection for segmentation and sparse sampling, which will help to improve the generalization ability of our proposed method.

\paragraph{Mobile and Wearable Implementations:} Our proposed method, MMTSA, is an efficient HAR approach that significantly outperforms existing state-of-the-art (SOTA) algorithms in terms of recognition performance while reducing computational load and inference latency. However, MMTSA has a large number of parameters, resulting in significant memory overhead, thus limiting its deployment on mobile and wearable devices. To address this limitation, we plan to explore model pruning and quantization techniques in future work, as well as extend sparse sampling strategies to spatial feature extraction, to further investigate the feasibility of deploying MMTSA on edge devices.

\section{Conclusion} 
In this paper, we present a novel architecture, the Multimodal Temporal Segment Attention Network (MMTSA), for efficient Human Activity Recognition (HAR) using multimodal sensor data from RGB cameras and Inertial Measurement Units (IMUs). MMTSA, leveraging Gramian Angular Field (GAF) as a data isomorphism mechanism, effectively represents the inherent properties of human activities in the IMU data. To further streamline the process, we applied a multimodal sparse sampling method, reducing data redundancy and enhancing computational efficiency. An additional inter-segment attention module was deployed for the efficient fusion of multimodal data.
Upon rigorous evaluation using three public datasets, MMTSA outperforms SOTA methods, demonstrating an 11.13\% cross-subject F1-score improvement on the MMAct dataset and significant reductions of 94\% in FLOPs and 82.6\% in inference latency in edge deployment.
We also discuss the effectiveness of each module of MMTSA and propose guidelines for efficient modeling and sparse sampling of IMU data for the HAR task. 
Based on the scalability and high efficiency of MMTSA, we give some future directions for further optimizing multimodal HAR methods in ubiquitous computing.

\section*{ACKNOWLEDGMENTS}
This work is supported by the Natural Science Foundation of China (NSFC) under Grant No. 62132010 and No. 62002198, Young Elite Scientists Sponsorship Program by CAST under Grant No.2021QNRC001, Tsinghua University Initiative Scientific Research Program, Beijing Key Lab of Networked Multimedia, Institute for Artificial Intelligence, Tsinghua University, and Beijing National Research Center for Information Science and Technology (BNRist).

\bibliographystyle{ACM-Reference-Format}
\bibliography{sample-base}

\end{document}